\pdfoutput=1

\documentclass[11pt]{article}

\usepackage[]{ACL2023}

\usepackage{times}
\usepackage{latexsym}

\usepackage[T1]{fontenc}

\usepackage[utf8]{inputenc}

\usepackage{microtype}

\usepackage{inconsolata}
\usepackage{amsmath}
\usepackage{amssymb}
\usepackage{inconsolata}
\usepackage{graphicx}
\usepackage{subfigure}
\usepackage{caption}
\usepackage{subcaption}
\usepackage{booktabs}
\usepackage{multirow}
\usepackage{arydshln}
\usepackage{amsmath}
\usepackage{cases}
\title{Retrieval-augmented Multilingual Knowledge Editing}

\author{%
Weixuan Wang, Barry Haddow, Alexandra Birch \\[1ex]
School of Informatics, University of Edinburgh\\
\small
weixuan.wang@ed.ac.uk, bhaddow@ed.ac.uk, a.birch@ed.ac.uk \\
\small
}

\begin{document}
\maketitle
\begin{abstract}

Knowledge represented in Large Language Models (LLMs) is quite often incorrect and can also become obsolete over time. Updating knowledge via fine-tuning is computationally resource-hungry and not reliable, and so knowledge editing (KE) has developed as an effective and economical alternative to inject new knowledge or to fix factual errors in LLMs. Although there has been considerable interest in this area, current KE research exclusively focuses on the monolingual setting, typically in English. However, what happens if the new knowledge is supplied in one language, but we would like to query the LLM in a different language? To address the problem of multilingual knowledge editing, we propose \textbf{Retrieval-augmented Multilingual Knowledge Editor (ReMaKE)} to update new knowledge in LLMs. ReMaKE can perform model-agnostic knowledge editing in multilingual settings. ReMaKE concatenates the new knowledge retrieved from a multilingual knowledge base with prompts. Our experimental results show that ReMaKE outperforms baseline knowledge editing methods by a significant margin and is the first KE method to work in a multilingual setting. We provide our multilingual knowledge editing dataset (MzsRE) in 12 languages, which along with code, and additional project information is available at \url{https://github.com/Vicky-Wil/ReMaKE}.

\end{abstract}

\section{Introduction}

Large Language Models (LLMs) are being used as sources of factual knowledge for search engines and other downstream tasks. Despite their considerable progress, facts induced by LLMs can be incorrect or become obsolete in a changing world. Pre-training from scratch or fine-tuning LLMs to adapt them to new knowledge is computationally expensive and not guaranteed to work. Knowledge editing (KE) techniques \citep{ft,ke,serac,ike} have been proposed as an effective and economic alternative to fine-tuning when specific facts need to be added or updated. KE could involve either updating the parameters of the model \citep{kn,mend,rome,memit,pemt} or adding extra components \citep{serac,ike,calinet,grace}. For example, KE could be used to correct the answer to this question \textit{``Who is the foreign secretary of the UK?''} from \textit{``James Cleverly''} (true until mid November 2023) to \textit{``David Cameron''}, who has recently been appointed to the post. 

In spite of considerable interest in this problem, current KE research focuses on the monolingual language setting, where both the injected knowledge and the subsequent queries to the LLM, are in English \citep{mend,rome,memit,serac,ike}. 
Companies serving a multilingual customer base need to consider the multilingual KE case, where KE is done in one language and this propagates to answers in all other languages. 
\citet{cross} explore the cross-lingual applicability of knowledge editing by evaluating KE methods on the English-Chinese cross-lingual scenario. However, their focus was to present the challenges and not to develop a working approach to KE in a multilingual setting. 

Inspired by in-context learning (ICL), in-context knowledge editing (IKE) uses prompts to edit factual knowledge. This is the only method which  has shown any positive results in the multilingual KE task setting~\citep{cross}. However, IKE requires explicit provision of new knowledge every time the LLM is used, confining its practicality and scalability in real-world applications. In addition, IKE suffers when irrelevant facts are provided in the prompt \citep{monitor} especially when a potentially large number of facts are edited.  


In this paper, we propose \textbf{Retrieval-augmented Multilingual Knowledge Editor (ReMaKE)} which combines multilingual retrieval from a knowledge base with in-context learning. This leverages the advantages of a knowledge bases' ability to scale and IKE's knowledge editing performance. 
ReMaKE concatenates the retrieved knowledge with the user query to create the prompt. The retrieval process is critical to alleviate the negative effects of unrelated information 
as the developed multilingual retriever can extract information highly relevant to user inputs, largely removing the contextual interference due to irrelevant facts. Furthermore, the retriever will only return knowledge if it is related to the query, greatly reducing the impact of KE on unedited knowledge. 

\begin{figure*}[htbp]
    \centering
    \includegraphics[scale=0.4]{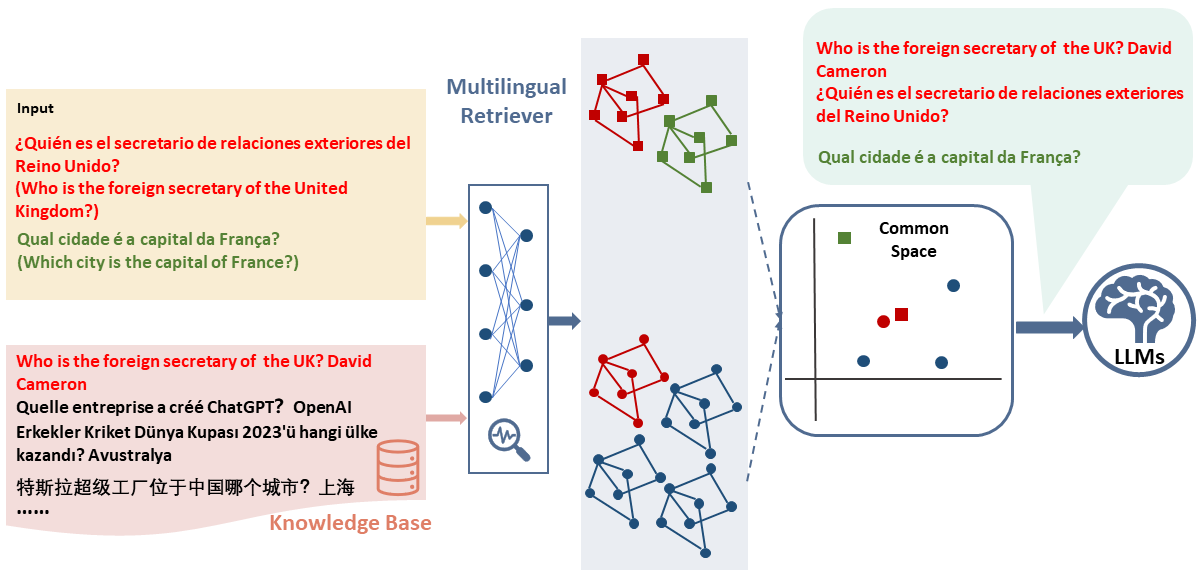}
    \caption{ReMaKE provides in-context knowledge to an LLM prompt when it is retrieved (red example where the edited knowledge is in English and user input is in Spanish) from a customer-defined multilingual knowledge base. When no edited knowledge is retrieved (green example) the prompt is passed to the LLM unchanged.} 
    \label{framework}

\end{figure*}

The generated prompts are designed to guide the LLMs in generating accurate responses associated with the injected knowledge. Figure~\ref{framework} shows the architecture of the proposed retrieval-augmented multilingual knowledge editor.

Our main contributions are listed below:

\begin{itemize}
\item[$\bullet$] \textbf{Multilingual knowledge editing}: To the best of our knowledge ReMaKE is the first multilingual knowledge editing framework. It can be \textbf{applied to any LLM} and it is \textbf{scalable}, extending to editing a large number of facts across different languages. 
\item[$\bullet$] \textbf{Evidence of ReMaKE's applicability}: We show that ReMaKE surpasses IKE across 12 languages showing large increases in average accuracy score from the smallest increase of +24.76 (for Czech) to the largest of +58.72 (for Russian) indicating that this approach is potentially ready for deployment at scale.


\item[$\bullet$] \textbf{Multilingual editing dataset}: We build a machine translated multilingual knowledge editing dataset (\textbf{MzsRE}) in 12 languages: English, Czech, German, Dutch, Spanish, French, Portugues, Russian, Thai, Turkish, Vietnamese and Chinese using the zsRE testset \citep{zsre}.
\end{itemize}

\section{Related Work}

\textbf{Knowledge editing:} Monolingual knowledge editing methods can be categorized into four main paradigms \citep{kesurvey,kesurvey1}: \textbf{Hypernetwork editors} \citep{ke,mend,remedi} re-frame knowledge editing as a learning-to-update problem with the help of gradient shift, which is predicted by extrinsic editors. While the scope extends beyond a single editing, the success rate of edits diminishes remarkably when more edits are executed simultaneously. \textbf{Locate-and-edit editors} \citep{kn,rome,memit,pemt} first locate the parameters related to factual knowledge and subsequently modify them. It is worth noting that this method requires an error-prone analytic step to identify parameters. It is model-specific and not efficient, as the locations are unique for each LLM. \textbf{Plug-in editors} \citep{ke,mend,remedi} add extra components to generate predictions about new knowledge without impacting on the parameters of the LLMs. Although this method has a low impact on unrelated inputs, it often cannot achieve precise editing. \textbf{Prompt-based editors} like IKE \citep{ike} use ICL to inject knowledge in the context of the prompt. Compared with other KE methods, IKE achieves a far stronger editing performance, together with far fewer side effects. However, IKE simply provides all new knowledge every time, limiting its practicality and scalability in real-world applications. 
All above mentioned editors are based on model-dependent monolingual methods, suffering from unreliable editing performance and low scalbility. Our proposed editor, ReMaKE, takes the problem and scales KE to the multilingual scenario covering many facts.

\textbf{Retrieval-augmented in-context learning:} 
ICL is a non-intrusive way to provide extra information for LLMs without impacting on the parameters, in which contexts are concatenated with an existing prompt to guide language generation. Furthermore, retrieval-augmented ICL is proposed to retrieve knowledge from an external datastore when needed. Off-the-shelf search engines are often used to enhance this process~\citep{retrieve0, retrieve1, retrieve2} finding semantically similar examples to the context to improve the performance of LLMs in a few-shot setting. In  cross-lingual scenarios, the search engines first uses an low-resource language input sample as a query to find the semantically most similar high-resource language sample in the corpus. The retrieved high-resource language sample together with the input sample are reformulated as prompts for LLMs. For instance, \citet{parc} retrieve semantically similar cross-lingual sentences as prompts to improve the performance of sentiment classification for low-resource languages. 

Whilst ICL can be used to support cross-lingual tasks, the problem of knowledge editing across language boundaries has not been explored. \citet{cross} show that KE in cross-lingual settings remains a challenge. 


\section{Retrieval-augmented Multilingual Knowledge Editing}
To design a scalable knowledge editing method across model and language boundaries, we propose a retrieval-augmented multilingual knowledge editor (\textbf{ReMaKE}). This enables knowledge to be edited in one language and subsequently queried in multiple languages. For example, one may edit the knowledge in English and test (by probing the edited facts)  in other languages. 

\textbf{ReMaKE} consists of two stages: multilingual knowledge retrieval and multilingual in-context editing.

\subsection{Multilingual Knowledge Retrieval}


We propose a simple multilingual retrieval model to search for the most relevant fact stored in the knowledge base for a query. As shown in Figure~\ref{framework}, the proposed retrieval model initially maps a query and knowledge base entries to a shared multilingual embedding space. We train a classifier on top of these embeddings to determine if a knowledge fact is semantically related to a query. The classifier is based on a sentence transformer (i.e. XLM-R), showing excellent performances on our test set (with retrieval accuracies >90\%). 

More specifically, we finetune the multilingual retrieval model $f_\theta$ with a binary classification head on the multilingual parallel dataset constructed by translating our English training dataset using Google Translate. We use the separator token \texttt{</s>} to concatenate the sentence $x$ and and its corresponding translation $I(x)$ to format the input, predicting whether they are semantically related (related: $f_\theta(x,I(x))=1$ vs. unrelated: $f_\theta(x,I(x))=0$). Negative examples are constructed by pairing unrelated sentences between languages. 

Once trained, the multilingual retriever $f_\theta$ takes the query $x_{l_1}$ in language ${l_1}$ and seeks the knowledge $k_{l_2}$ in language ${l_2}$. From new knowledge base $K_{l_2} = \{k_{l_2}^0,..,k_{l_2}^i,...,k_{l_2}^K\}$, the retriever $f_\theta$ iterates across each knowledge item for the query and returns the most related knowledge or empty $R(x_{l_1})$:

 \begin{gather} 
    k_{l_2} = R(x_{l_1}) =  \begin{cases}k_{l_2}^{i^*} & f_\theta(x_{l_1},k^{i^*}_{l_2}) = 1\\ None & f_\theta(x_{l_1},k^{i^*}_{l_2}) = 0\end{cases}
    \label{retriever}
\end{gather}
where $i^* = argmax_iP(f_\theta(x_{l_1},k_{l_2}^{i})=1|i)$ is the index that maximizes the probability $P(f_\theta(x_{l_1},k_{l_2}^{i})=1|i)$.


It should be noted that ReMaKE can be extended to accommodate a more efficient and performant Information Retrieval model for real world deployment. We leave this extension as one of our future endeavours.

\begin{figure}[htbp]
    \centering
    \includegraphics[scale=0.5]{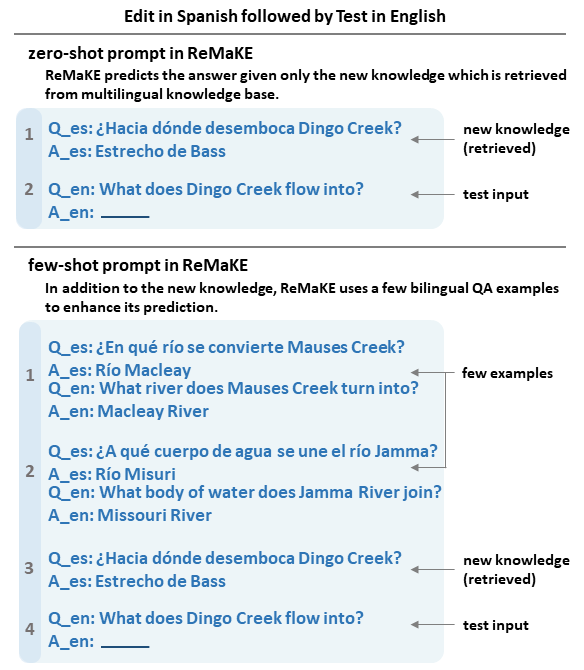}
    \caption{\textbf{Zero-shot and few-shot editing with ReMaKE.} The panels above show two methods for performing multilingual KE, in which an fact is edited in Spanish subsequently evaluated using an English question. ``Q\_en, A\_en'' and ``Q\_es, A\_es'' are the QA pairs in English and Spanish.
    }
    \label{data-fig}

\end{figure}

\subsection{Multilingual In-context Editing}
ReMaKE performs zero-shot and few-shot editing. In \textbf{zero-shot editing}, the retrieved result (``\textit{new knowledge}'' in the Figure~\ref{data-fig}) is concatenated with the user's input (``\textit{test input}'' in  Figure~\ref{data-fig}) to form a prompt (``\textit{zero-shot prompt}'' in Figure~\ref{data-fig}) to predict the output $P(y_{l_1}|x_{l_1},k_{l_2}^{i^*})$.

In \textbf{few-shot editing}, bilingual examples $S = \{(s^{1}_{l_1},s^1_{l_2}),...,(s^q_{l_1},s^q_{l_2})\}$ are added before the new knowledge and the test input, where $s^{j}_{l_1}$ and $s^{j}_{l_1}$ is the same statement in language $l_1$ and $l_2$, corresponding to the ``\textit{Q\_es: ... A\_es: ...}'' and ``\textit{Q\_en: ... A\_en: ...}(''\textit{few examples}'' in Figure~\ref{data-fig}). In few-shot editing, we concatenate ``\textit{few examples}'', ``\textit{new knowledge}'', ``\textit{test input}'' as the prompt (``\textit{few-shot prompt}'' in Figure~\ref{data-fig}). The goal of predicting an edited fact is $P(y_{l_1}|x_{l_1},k_{l_2}^{i^*},S)$. For the few-shot setting, we follow  \citet{ike} in selecting examples with an unsupervised method from the training corpus based on their cosine similarity to the inputs with using all-MiniLM-L6-v2\footnote{https://huggingface.co/sentence-transformers/all-MiniLM-L6-v2}. The selected examples are included in the context to perform in-context learning.

\section{Metrics, Data and Model}

\subsection{Metrics}

\label{metrics}
Following ~\citet {cross}, we evaluate multilingual knowledge editing with the following four metrics: (1) \textit{Reliability} evaluates the average accuracy of an LLM on all edited instances.  (2) \textit{Generality} measures the average accuracy of an LLM for the paraphrased inputs for all edited instances. It indicates ReMaKE's effectiveness under the prompting frame bias ~\citep{monitor} induced by paraphrasing. (3) \textit{Locality} assesses the average accuracy of an LLM in response to queries on irrelevant semantics after knowledge editing. It tests the knowledge editors ability to update only the desired knowledge, without affecting other knowledge in the model. (4) \textit{Portability} estimates the average accuracy of an LLM for questions requireing reasoning after knowledge editing. Reasoning questions are constructed to test an LLM's ability to provide answers requiring it to reason. Portability can indicate if KE effectively adapts an LLM's knowledge to support reasoning.  

\subsection{Data Construction}

Zero-Shot Relation Extraction (zsRE) \citep{zsre} is a monolingual question-answering test set containing 1,038 samples widely used in the knowledge editing task. There is a question-answer pair for each fact where the answer is an alternative counterfactual prediction \citep{ke}. The counterfactual answer is expected to be generated by the post-edited LLMs. We translate the QA pairs and store them in the  knowledge base. Additionally, a paraphrased question, an unrelated question, and a portability question are provided to evaluate the generality, the locality, and the portability of the editing. 
We translate the zsRE from English to ten languages: Czech, German, Dutch, Spanish, French, Portuguese, Russian, Thai, Turkish, and Vietnamese with Google Translate and use the Chinese zsRE test (which was also machine translated from zsRE) set from \citep{cross} to construct the multilingual zsRE test set (MzsRE). As all new knowledge is stored in the knowledge base, each sample should be unique and because there are some different answers for the same question in zsRE, we de-duplicate MzsRE to 743 items to avoid conflicting entries in the knowledge base. Table~\ref{statistics} (Appendix~\ref{statistics-appendix}) lists the statistics of MzsRE. Figure~\ref{data-fig} illustrates a sample case of multilingual KE. More specifically, we show an example of ES (edit) $\rightarrow$ EN (test) knowledge editing for four metrics in Table~\ref{data-example}. 

\begin{table*}[h] \small
\centering
\begin{tabular}{llll}
\toprule
 & Question & Answer  & Ground Truth    \\ \hline
New Knowledge & ¿Qué ciudad fue el lugar de nacimiento de Henning Löhlein? & Munich    &   Bonn                 \\ \hline
Reliability   & Which city was the birthplace of Henning Löhlein?          & Munich     & Bonn                   \\
Generality    & In which city is Henning Löhlein born?                        & Munich      & Bonn                  \\
Locality      & Who is the lead singer of collective soul?         & Ed Roland & Ed Roland\\
Portability   & In which German state was Henning Löhlein born?            & Bavaria       & North Rhine              \\
\bottomrule
\end{tabular}
\caption{\label{data-example}An example of ES (edit) $\rightarrow$ EN (test) knowledge editing for four metrics. ``Answer'' represents the counterfactual post-edited knowledge which is needed to predict, and ``Ground Truth'' is the factual knowledge. }
\end{table*}

\begin{table*}[htbp] \scriptsize
\centering
\begin{tabular}{llllllllllllll}
\toprule
\multirow{2}{*}{\textbf{Metrics}}                     & \multirow{2}{*}{\textbf{Edit on EN}}     & \multicolumn{12}{c}{\textbf{Test on}} \\ \cdashline{3-14}[1pt/1pt]
 & & \textbf{EN}     & \textbf{CS}     & \textbf{DE}     & \textbf{NL}     & \textbf{ES}     & \textbf{FR}     & \textbf{PT}     & \textbf{RU}     & \textbf{TH}     & \textbf{TR}     & \textbf{VI}     & \textbf{ZH}     \\ \hline
\multirow{7}{*}{\textbf{Reliability}} & LLaMA & 1.08	& 0.13	& 0.54	& 0.27	& 0.13	& 0.27	& 0.27	& 0.40	& 0.27	& 0.54	& 0.13	& 1.21 \\ \cdashline{2-14}[1pt/1pt]
& \textbf{SERAC}    &  91.25           & 0.00           & 0.00           & 0.00           & 0.00           & 0.00           & 0.00           & 0.00           & 0.00           & 0.00           & 0.00           & 0.00            \\
& \textbf{ROME}     & 68.91 &  10.77   &  16.02  &  15.48   &   12.25  &  10.09 &  12.11   &  0.13   &   0.13   &  1.21   &  4.31    &  1.21  \\ 
& \textbf{IKE}        &  100.0 & 50.34          & 51.49          & 44.26          & 36.45          & 43.39          & 38.09          & 3.86           & 3.18           & 39.44          & 40.02          & 6.36         \\ \cdashline{2-14}[1pt/1pt]
& \textbf{ReMaKE}-zero  & 96.37           & 61.10          & 64.87          & 54.91          & 52.62          & 53.43          & 54.51          & 27.73          & 5.92           & 45.22          & 48.32          & 25.44          \\
& \textbf{ReMaKE}-few-mono & 100.0 & 56.26 & 57.87 & 49.93 & 43.47 & 48.32 & 45.49 & 19.78 & 5.65 & 43.47 & 41.72 & 17.63 \\
& \textbf{ReMaKE}-few-bi   & \textbf{100.0} & \textbf{75.10} & \textbf{81.70} & \textbf{72.68} & \textbf{68.10} & \textbf{73.35} & \textbf{71.20} & \textbf{62.58} & \textbf{32.44} & \textbf{70.79} & \textbf{68.37} & \textbf{54.78} \\ \hline
 
\multirow{7}{*}{\textbf{Generality}}& LLaMA & 0.94	& 0.13	& 0.94	& 0.40	& 0.13	& 0.13	& 0.13	& 0.27	& 0.13	& 0.13	& 0.13	& 1.48 \\ \cdashline{2-14}[1pt/1pt]
& \textbf{SERAC}        &  26.78           & 0.00           & 0.00           & 0.00           & 0.00           & 0.00           & 0.00           & 0.00           & 0.00           & 0.00           & 0.00           & 0.00         \\
& \textbf{ROME}     & 56.53  &  10.90  &  14.40  &   11.96 &  11.71    &   8.34 &  9.56  &  0.13  &  0.00    &    1.48  &  4.17   & 0.81 \\  
& \textbf{IKE}          &  \textbf{98.65}  & 49.76          & 51.49          & 43.88          & 35.39          & 42.91          & 37.61          & 3.38           & 3.18           & 39.15          & 39.34          & 5.98       \\ \cdashline{2-14}[1pt/1pt]
& \textbf{ReMaKE}-zero & 86.81           & 57.60          & 62.85          & 53.16          & 50.34          & 50.74          & 51.01          & 24.50          & 6.06           & 42.66          & 46.03          & 23.01          \\
& \textbf{ReMaKE}-few-mono & 98.25  & 55.59 & 57.34 & 48.59 & 43.61 & 47.64 & 44.68 & 18.57 & 5.52 & 42.4 & 41.18 & 17.23 \\
& \textbf{ReMaKE}-few-bi  & 98.25  & \textbf{73.76} & \textbf{80.62} & \textbf{71.60} & \textbf{67.97} & \textbf{71.60} & \textbf{70.66} & \textbf{62.45} & \textbf{32.97} & \textbf{70.12} & \textbf{67.83} & \textbf{53.57} \\ \hline
 
\multirow{6}{*}{\textbf{Locality}}    & \textbf{SERAC}        &  99.46           & 100.0         & 100.0         & 100.0         & 100.0         & 100.0         & 100.0         & 100.0         & 100.0         & 100.0         & 100.0         & 99.87     \\
& \textbf{ROME}     & 92.87 &  84.25  &  87.48   &  87.08   &   88.83  &  88.56 &  86.54  &  84.39  &   97.31  &   87.21   &  95.02    &  91.92  \\ 
& \textbf{IKE}          &   38.48           & 0.39           & 5.69           & 1.54           & 1.74           & 0.48           & 0.48           & 0.19           & 1.35           & 0.96           & 0.96           & 0.96         \\ \cdashline{2-14}[1pt/1pt]
& \textbf{ReMaKE}-zero & \textbf{99.46}           & \textbf{98.65}          & \textbf{99.73}          & \textbf{99.87}          & \textbf{98.52  }        & \textbf{99.06 }         & \textbf{99.19 }         & \textbf{97.58 }         & \textbf{95.29 }         & \textbf{97.17}          & \textbf{97.71}          & \textbf{94.48}          \\
& \textbf{ReMaKE}-few-mono & 99.46 & 98.38 & 99.6 & 99.73 & 98.52 & 99.06 & 99.19 & 97.58 & 95.29 & 97.04 & 97.71 & 94.48 \\
& \textbf{ReMaKE}-few-bi & 99.46           & 98.25          & 99.60          & 99.73          & 98.25          & 98.92          & 99.19          & 97.44          & 95.29          & 97.04          & 97.71          & 93.94          \\ \hline

\multirow{7}{*}{\textbf{Portability}} & LLaMA & 8.48	& 2.29	& 3.50	& 2.83	& 3.90	& 2.29	& 3.10	& 0.54	& 0.27	& 0.94	& 1.88	& 1.08 \\ \cdashline{2-14}[1pt/1pt]
& \textbf{SERAC}        &  0.00            & 0.00           & 0.00           & 0.00           & 0.00           & 0.00           & 0.00           & 0.00           & 0.00           & 0.00           & 0.00           & 0.00      \\ 
& \textbf{ROME}     &  0.00 &   0.00  &   0.00   &    0.00     &   0.00   &   0.00  &  0.00   &   0.00   &    0.00  &  0.00   &  0.00   &  0.00  \\  
& \textbf{IKE}          &    17.26           & 1.54           & 4.63           & 3.28           & 1.93           & 2.51           & 2.89           & 0.10           & 0.10           & 0.87           & 1.74           & 0.10      \\
\cdashline{2-14}[1pt/1pt]
& \textbf{ReMaKE}-zero & \textbf{34.59}  & \textbf{12.11} & \textbf{18.30} & \textbf{13.73} & \textbf{11.71} & \textbf{12.25} & \textbf{12.92} & \textbf{3.50}  & \textbf{0.27}  & \textbf{5.38}  & \textbf{9.83}  & 3.63           \\
& \textbf{ReMaKE}-few-mono & 31.49  & 6.46 & 11.57 & 9.69 & 10.23 & 8.48 & 10.23 & 2.02 & 0.13 & 4.04 & 5.79 & 2.42 \\
& \textbf{ReMaKE}-few-bi & 31.49           & 7.67           & 11.31          & 9.02           & 8.61           & 8.08           & 9.83           & 5.79           & 0.67           & 3.50           & 5.25           & \textbf{5.92}           \\
\bottomrule              
\end{tabular}
\caption{\label{en2xx-em}Exact Match (EM) results on the LLaMA backbone obtained from testing in English, Czech, German, Dutch, Spanish, French, Portuguese, Russian, Thai, Turkish, Vietnamese and Chinese after performing KE on knowledge in English. ``ReMaKE-few-bi'' means the proposed knowledge editing method leveraging few-shot learning based on 16 bilingual examples concatenated in the context. ``ReMaKE-few-mono'' and ``IKE'' use 16 monolingual (English) in the context. ``LLaMA'' are the results of pre-editing.}
\end{table*}

\begin{table*}[htbp] \scriptsize
\centering
\begin{tabular}{lllllllllllll}
\toprule

\multirow{2}{*}{\textbf{Metrics}}                     & \multirow{2}{*}{\textbf{Test on EN}}     & \multicolumn{11}{c}{\textbf{Edit on}} \\ \cdashline{3-13}[1pt/1pt]
 &    & \textbf{CS}     & \textbf{DE}     & \textbf{NL}     & \textbf{ES}     & \textbf{FR}     & \textbf{PT}     & \textbf{RU}     & \textbf{TH}     & \textbf{TR}     & \textbf{VI}     & \textbf{ZH}     \\ \hline
\multirow{6}{*}{\textbf{Reliability}} 
& LLaMA &   1.08   & 1.08    &  1.08  &  1.08   & 1.08     & 1.08        &  1.08    & 1.08  &  1.08    & 1.08     & 1.08 \\  \cdashline{2-13}[1pt/1pt]
& \textbf{ROME}     &  16.96   &  37.55  &  35.13    &   32.84   &  32.57  &   31.49  &   1.35  &   0.00   &  3.90  &  4.85 &  0.94 \\  
& \textbf{IKE}          &  57.67          & 55.45          & 50.05          & 40.21          & 46.38          & 43.20          & 52.36          & 2.03           & 40.31          & 41.85          & 20.54   \\ \cdashline{2-13}[1pt/1pt]
& \textbf{ReMaKE}-zero & 69.18          & 65.68          & 60.97          & 62.31          & 66.22          & 59.76          & 59.49          & 9.96           & 50.47          & 51.14          & 44.68          \\
& \textbf{ReMaKE}-few-mono & 62.72 & 61.37 & 55.05 & 45.76 & 56.8  & 48.72 & 60.16 & 2.83  & 49.93 & 50.2  & 41.86 \\
& \textbf{ReMaKE}-few-bi & \textbf{87.89} & \textbf{90.17} & \textbf{87.21} & \textbf{86.41} & \textbf{86.41} & \textbf{86.68} & \textbf{82.91} & \textbf{49.26} & \textbf{82.10} & \textbf{84.66} & \textbf{72.14} \\\hline
 
\multirow{6}{*}{\textbf{Generality}} & LLaMA &  0.94    &   0.94  &  0.94  &  0.94   &  0.94    &   0.94      &   0.94   & 0.94  &  0.94    &  0.94    & 0.94 \\  \cdashline{2-13}[1pt/1pt]
& \textbf{ROME}     &   16.02   &   35.53  &   32.30    &   30.15   &  30.15 &  26.78  &  1.62  &  0.00 &   3.90 &  3.50 & 0.81  \\ 
& \textbf{IKE}          &   56.41          & 54.39          & 49.08          & 39.25          & 45.03          & 42.91          & 49.47          & 2.03           & 39.15          & 40.89          & 20.64      \\ \cdashline{2-13}[1pt/1pt]
& \textbf{ReMaKE}-zero & 63.26          & 62.05          & 54.37          & 53.84          & 61.10          & 54.64          & 53.84          & 9.29           & 47.51          & 46.70          & 40.38          \\
& \textbf{ReMaKE}-few-mono  & 61.1  & 60.43 & 53.3  & 45.09 & 56.66 & 48.32 & 57.6  & 2.56  & 48.86 & 49.66 & 42.13 \\
& \textbf{ReMaKE}-few-bi & \textbf{87.75} & \textbf{88.96} & \textbf{86.00} & \textbf{84.66} & \textbf{84.93} & \textbf{85.60} & \textbf{82.10} & \textbf{48.99} & \textbf{80.35} & \textbf{84.25} & \textbf{69.99} \\ \hline
 
\multirow{5}{*}{\textbf{Locality}}   & 
 \textbf{ROME}     & 84.66   &   87.89    &  86.94  &  88.83   &  87.89   &  85.33  &  82.37  &  97.04 & 90.04  &  93.54  &  92.73 \\ 
& \textbf{IKE}          &    1.25           & 1.16           & 1.16           & 1.06           & 1.16           & 1.25           & 0.87           & 0.10           & 1.16           & 1.06           & 0.96         \\ \cdashline{2-13}[1pt/1pt]
& \textbf{ReMaKE}-zero & 98.92          & \textbf{99.06 }         & \textbf{99.46}          & \textbf{98.52 }         & \textbf{98.92 }         & \textbf{98.92  }        & \textbf{98.12 }         & \textbf{97.58 }         & 97.31          & \textbf{98.79 }         & \textbf{99.33 }         \\
& \textbf{ReMaKE}-few-mono  & \textbf{99.06} & 98.52 & 98.92 & 98.52 & 98.65 & 98.92 & 98.12 & 97.04 & \textbf{97.44} & 98.79 & 99.06 \\
& \textbf{ReMaKE}-few-bi  & 98.79          & 98.38          & 98.79          & 98.52          & 98.79          & 98.92          & 97.98          & 97.17          & 97.31          & 98.79          & 99.19          \\ \hline

\multirow{6}{*}{\textbf{Portability}} & LLaMA &   8.48   &  8.48   & 8.48   &  8.48   &  8.48    &   8.48      &    8.48  &  8.48 & 8.48     &   8.48   &8.48 \\  \cdashline{2-13}[1pt/1pt]
& \textbf{ROME}     &  0.00   &  0.00   &   0.00  &  0.00    &  0.00    &   0.00  &  0.00   &   0.00    &  0.00  &  0.00 &  0.00  \\ 
& \textbf{IKE}          &   5.69           & 7.43           & 5.88           & 5.50           & 2.89           & 5.11           & 7.62           & 0.10           & 2.12           & 4.34           & 1.06      \\ \cdashline{2-13}[1pt/1pt]
& \textbf{ReMaKE}-zero & \textbf{25.71} & \textbf{27.99} & \textbf{26.65} & \textbf{25.44} & \textbf{24.63} & \textbf{26.11} & \textbf{20.86} & \textbf{11.57} & \textbf{22.48} & \textbf{24.09} & \textbf{19.65} \\
& \textbf{ReMaKE}-few-mono  & 19.38 & 23.42 & 19.65 & 18.98 & 20.59 & 20.59 & 16.55 & 3.36  & 13.73 & 14.27 & 16.29 \\
& \textbf{ReMaKE}-few-bi  & 17.50          & 21.53          & 19.92          & 18.71          & 19.11          & 19.25          & 13.19          & 13.06          & 17.77          & 19.65          & 16.15           \\
\bottomrule              
\end{tabular}
\caption{\label{xx2en-em}EM (Exact Match) results on the LLaMA backbone obtained from testing in English after performing KE on knowledge in Czech, German, Dutch, Spanish, French, Portuguese, Russian, Thai, Turkish, Vietnamese and Chinese. ``ReMaKE-few-bi'' means the proposed knowledge editing method leveraging few-shot learning based on 16 bilingual examples concatenated in the context. ``ReMaKE-few-mono'' and ``IKE'' use 16 monolingual (editing language) in the context. ``LLaMA'' are the results of pre-editing.}
\end{table*}

\subsection{Base LLMs}

Two representative multilingual LLMs are selected as backbones for us to test various KE methods in the experiments: LLaMA2-7b and BLOOMz-7b1-mt, where LLaMA2-7b\footnote{https://huggingface.co/meta-LLaMA/LLaMA-2-7b-hf} \citep{LLaMA} is a foundation model and BLOOMZ-7b1-mt\footnote{https://huggingface.co/bigscience/bloomz-7b1-mt} \citep{bloomz} is an instruction-finetuned model. We translate a random sample of 10,000 instances from the zsRE training dataset into the other 11 languages and finetune an XLM-RoBERTa-base\footnote{https://huggingface.co/xlm-roberta-base} \citep{xlmr} on this multilingual dataset to develop our multilingual retriever.   

\subsection{Implementation Details}

All experiments are conducted on a single NVIDIA A-100 GPU (80G). The implementation is based on the EasyEdit \citep{easyedit} framework.

\subsection{Baseline}

We choose three KE baselines for the experiment which have shown the best performance in \citet{cross}. \textbf{IKE} is a baseline to apply in-context learning to knowledge editing, where the prompt consists of one explicit piece of knowledge in the editing language, one query in the testing language, and a certain number of examples in the editing language (16 in this case, following the setting in \citet{cross}). We also test a memory-based KE method \textbf{SERAC} with a memory size $K$ ($K=10$ is the default parameter in the \citet{serac}), which uses a classifier and a counterfactual model (another LLM) to generate a prediction in the testing language based on the new knowledge in editing knowledge. The classifier and counterfactual model are pre-trained on the monolingual dataset in the editing language. The parameters of the LLM for both methods mentioned above are frozen. To compare the effect of parameter-updating KE, We evaluate the \textbf{ROME} method, which locates the knowledge in the editing language first and subsequently performs editing. After updating the parameters, we evaluate the performance with a query in testing language. In the experiments, all baselines use their original proposed default parameters and LLaMA2-7b as the backbone.

\section{Experimental Results}

In order to discuss the results we refer to experiments, for example, ``ES (edit) $\rightarrow$ EN (test)'' where Spanish is the language in which we edited the knowledge and English is the language in which we tested the knowledge, as shown in the Figure~\ref{data-fig}. All results of this section are evaluated with setting the knowledge base as the whole test set. 

\subsection{English-based Multilingual Knowledge Editing}

In this subsection experiments are focused on English as either the editing or testing language. The evaluation results of LLMs on the LLaMA backbone in 12 languages after editing in English (aka ``EN (edit) $\rightarrow$ ALL (test)'') are shown in Table~\ref{en2xx-em} (based on Exact Match (EM)) and Table~\ref{en2xx-f1} (based on F1 score). Experimental results on the LLaMA backbone obtained from ``ALL (edit) $\rightarrow$ EN (test)'' are shown in Table~\ref{xx2en-em} and Table~\ref{xx2en-f1}. We compare ReMaKE with LLaMA under the zero-shot (``ReMaKE-zero''), monolingual few-shot setting (``ReMaKE-few-mono''), and bilingual few-shot settings (``ReMaKE-few-bi'') with three baseline methods and pre-editing results (``LLaMA''). 

As shown in Table~\ref{en2xx-em}, current KE approaches which work reasonably well in the monolingual case (See Reliability for SERAC, IKE, and ROME for ``EN (edit) $\rightarrow$ EN (test)'') either do not work at all or perform poorly in a multilingual setting. The pre-editing results of ``LLaMA'' fails (less than 2\%) because the knowledge editing test examples are counterfactual. SERAC scores all zeros in the multilingual case except for the Locality metric (wrt irrelevant queries) and ROME performs similarly poorly. IKE shows 100\% accuracy for monolinugal KE, and performs considerably better than ROME and SERAC. We chose to base ReMaKE on in-context learning due to the promising results of monolingual IKE. ReMaKE reveals a significant improvement over IKE in multilingual language conditions. ReMAKE, although fundamentally similar to IKE, provides bilingual few-shot examples and an additional means to filter out irrelevant queries (by returning null knowledge), leading to significant improvements in all four metrics. Furthermore, the accurate retriever ensures the scalability and the precision of editing.



For Reliability (average accuracy), the range of improvement ReMaKE has over baselines ranges from +24.76 (for Czech) to +58.72 (for Russian). 
Take ``EN (edit) $\rightarrow$ ES (test)'' as an example, SERAC has the worst reliability score (0.00) as the counterfactual model (used for generate prediction about new knowledge) in SERAC is monolingual, and IKE and ROME have reliability scores of 36.45 and 12.25, respectively. ReMaKE-zero achieves a reliability score of 52.62 instead. When scaled up to a few-shot setting, ReMaKE-few-mono drops to 43.47 due to the negative influence of the monolingual context, but adding bilingual examples to the context makes ReMaKE-few-bi the most capable KE with a reliability score 68.10. 

With regard to the results of ``ALL (edit) $\rightarrow$ EN (test)'' \footnote{A counterfactual model is required for each language for SERAC, leading to significant computation overhead. It is not included in this experiment to this end.}, ReMaKE-few-bi achieves the highest scores for the reliability and generality metrics. It records a reliability score 86.41 in ``ES (edit) $\rightarrow$ EN (test)''. The proposed ReMaKE overall excels in reliability and generality scores. 

There are some discrepancies in the KE across languages for backbones, reflecting the different capabilities of multilingual LLMs. After editing knowledge expressed in English, ReMaKE-few-bi attains a reliability score of 81.70 (``EN (edit) $\rightarrow$ DE (test)'') when testing the LLM on DE -- the highest across all languages. The lowest reliability score of 32.44 is from (``EN (edit) $\rightarrow$ TH (test)''), indicating the effect of KE on LLMs is sensitive to language settings. A similar phenomenon can be observed for the same KE method (ReMaKE-few) on a different backbone LLM (i.e., BLOOMZ in Appendix~\ref{em-bloomz}). We guess reason behind this sensitivity is caused by the distribution of training data in the LLM, but we can observe that a high-resource language and a powerful LLM are preferable choices. Even though ReMaKE appears sensitive to language settings and backbone LLMs, it demonstrates consistently significant effects on all languages and backbone LLMs in the experiment.

After knowledge editing, the locality of the LLMs can be significantly influenced, as shown in Tables \ref{en2xx-em}-\ref{xx2en-em}. For the locality, it is calculated by comparing pre-edit and post-edit predictions to show that irrelevant input is not affected, although the pre-edit answers are sometimes wrong. IKE performs poorly in locality, with most of the scores (measured in EM) below 1. It can be observed that ReMaKE can achieve consistently high locality scores across language settings and backbone LLMs due to its mitigation of contextual interference. 

\label{results-portability}
All KE methods record very low portability scores due to their ineffectiveness in impacting LLMs' reasoning capability. To understand the mechanism responsible for the reasoning capability of an LLM remains a challenge. It is worth noting that ReMaKE-zero outperforms all other KE methods in portability scores, largely attributed to the non-intrusive nature of ReMaKE on the LLMs' reasoning capability. The reason ReMaKE-zero performs better than its few-shot sibling, ReMaKE-few, is associated with its lower degree of contextual interference posed by the examples to an LLM.

\begin{figure}[h]
\centering     
\includegraphics[scale=0.26]{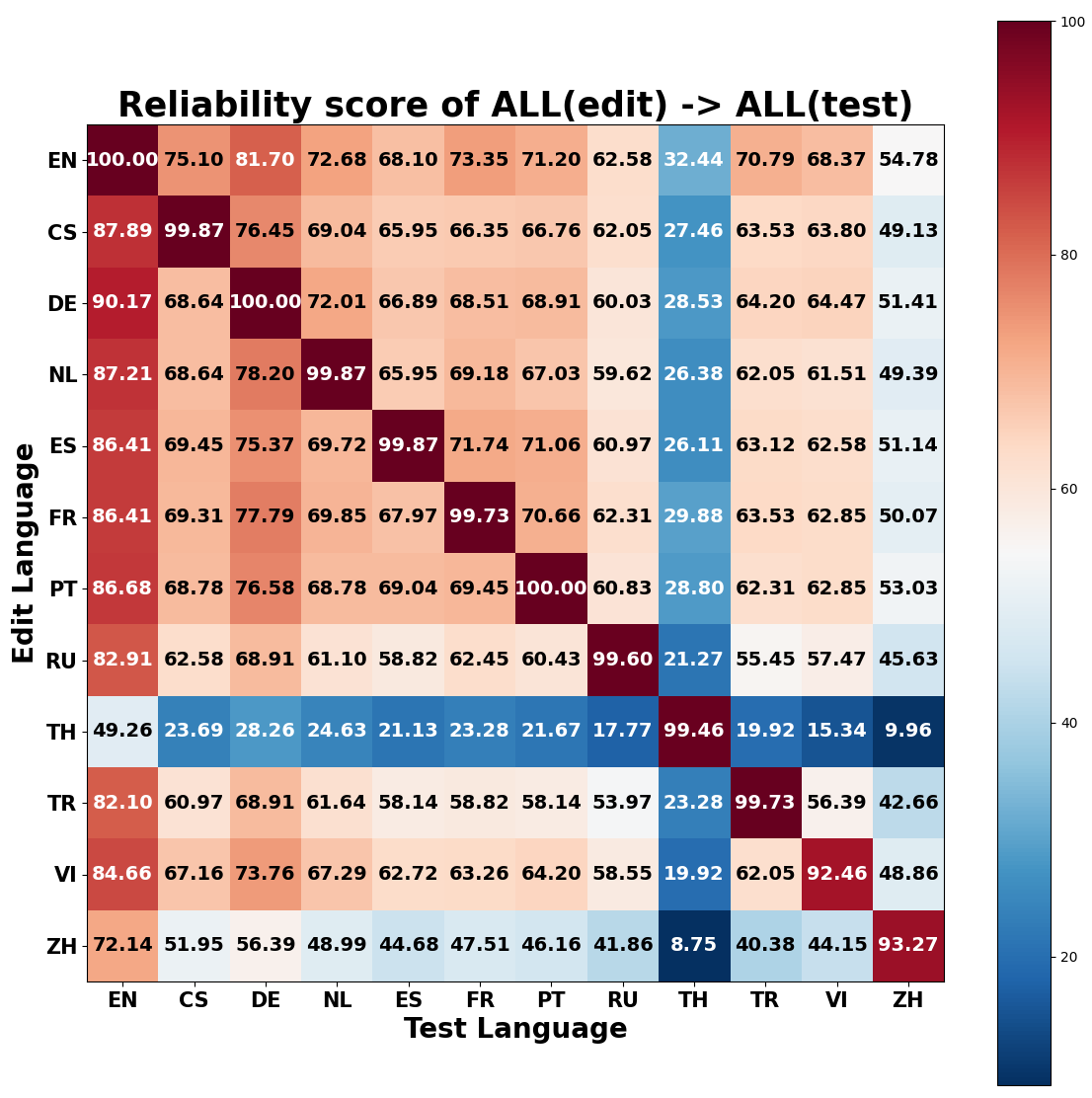}
\caption{Reliability score of multilingual knowledge editing.}     
\label{acc-xx2yy} 
\end{figure}

\begin{figure}[h]
    \centering
    \includegraphics[scale=0.26]{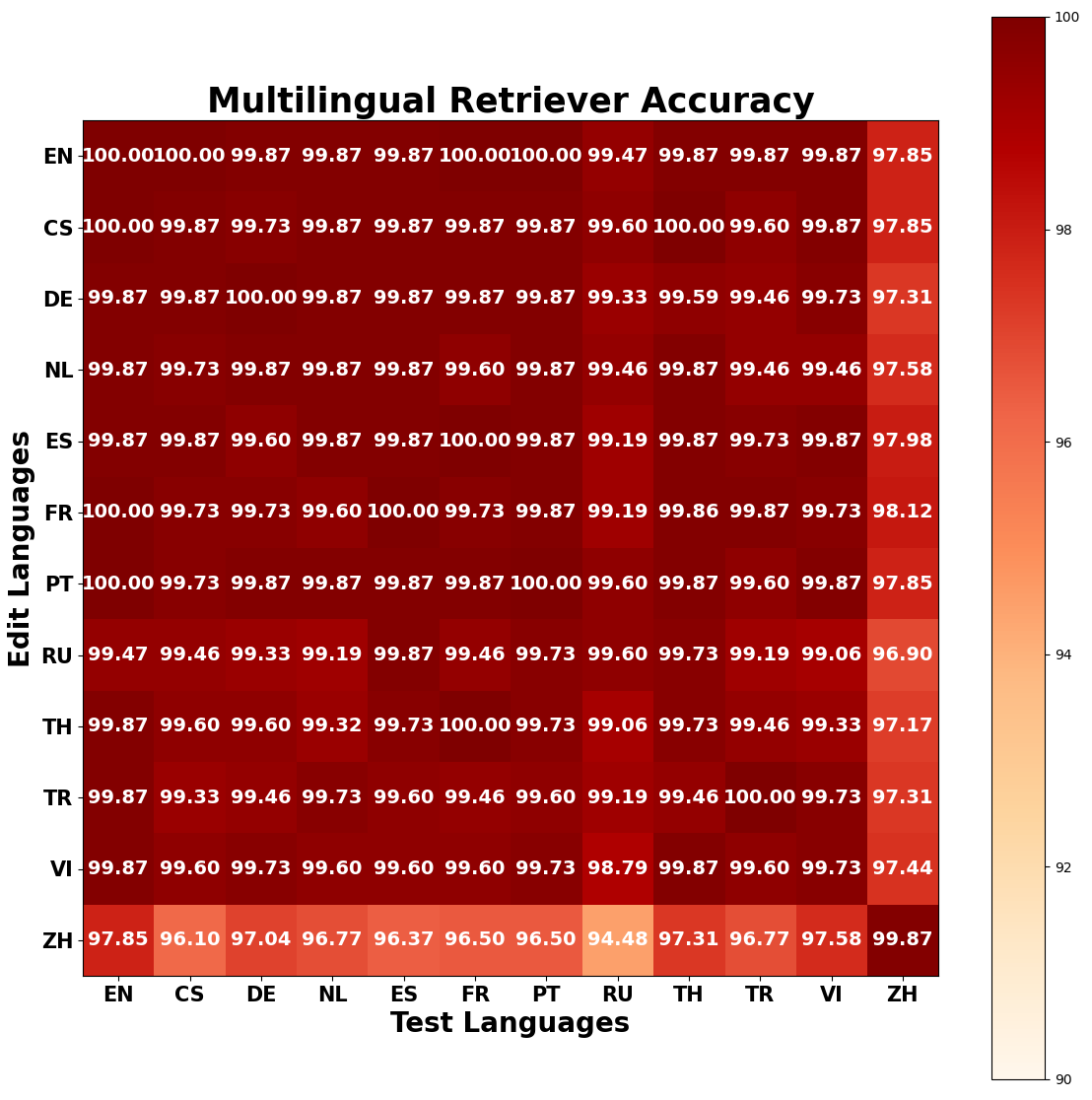}
    \caption{The retriever accuracy among 12 languages evaluated on the MzsRE dataset reliability metric.}
    \label{ret-acc-heat}

\end{figure}

\begin{figure}[h]
    \centering
    \includegraphics[scale=0.31]{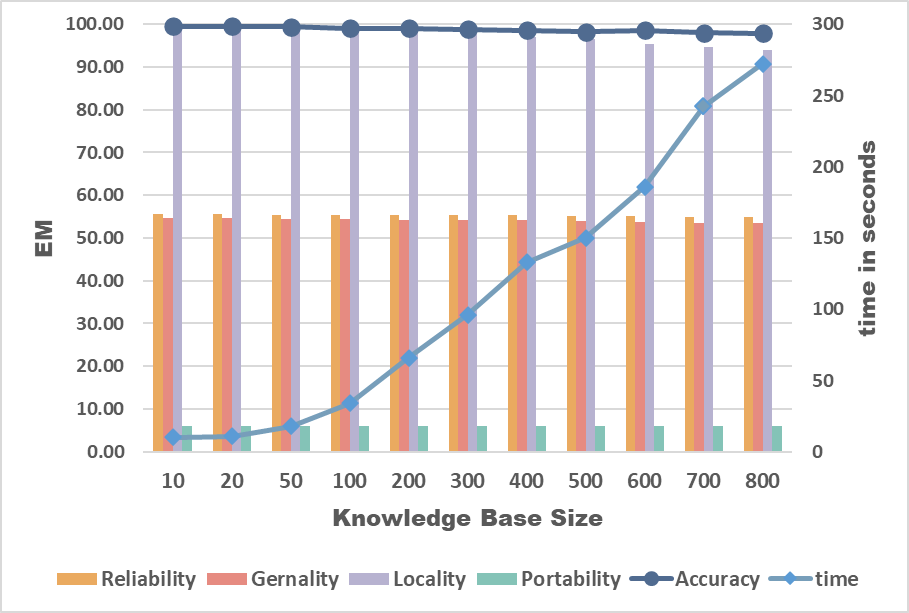}
    \caption{The results of an ablation study of effects of the size of the knowledge base on a variety of benchmark metrics when performing ReMaKE-few-bi editing on ``EN (edit) $\rightarrow$ ZH (test)'' on the LLaMA backbone. The time consumed is evaluated for the whole test.}
    \label{memory-size}

\end{figure}

\subsection{Multilingual Knowledge Editing between All Languages}

In this subsection, we extend the assessments to all involved twelve languages (``ALL (edit) $\rightarrow$ ALL (test)''). The results of reliability, generality, locality, and portability based on EM are illustrated as heat maps in Figure~\ref{acc-xx2yy} and Figure~\ref{acc-xx2yy-1}. The discrepancies presented in the reliability and generality scores between a certain language group (i.e., ZH, RU, TH, and TR) and the rest of the language groups are significant. It appears a natural segregation exists between these special languages and the rest of the languages, probably due to their linguistic characteristics and the language distribution in the training dataset.

Moreover, the portability scores captured in Figure~\ref{por-xx2yy} are below 10, which are much lower than those shown in English-pivoted multilingual KE (Tables~\ref{en2xx-em}-~\ref{xx2en-em}). It is more challenging for multilingual KE methods to influence the reasoning capability of an LLM when English is not in the loop. It is suggested that English be used as a pivot for multilingual KE to this end.

\subsection{Retriever Accuracy}

We further investigate the accuracy of the multilingual retriever of ReMaKE using sampled sentence pairs in the MzsRE dataset. The results are captured in Figure~\ref{ret-acc-heat}. The retriever achieves an accuracy of over 90\% for all languages. The sub-optimal retrieval accuracies for some languages (i.e., Chinese, Russian) may contribute to the sub-optimal performance of multilingual KE results in these languages.

\section{Analysis and Discussion}

\subsection{Ablation Study on Few-shot Learning}

\begin{figure}[htbp]
\centering       
\includegraphics[scale=0.55]{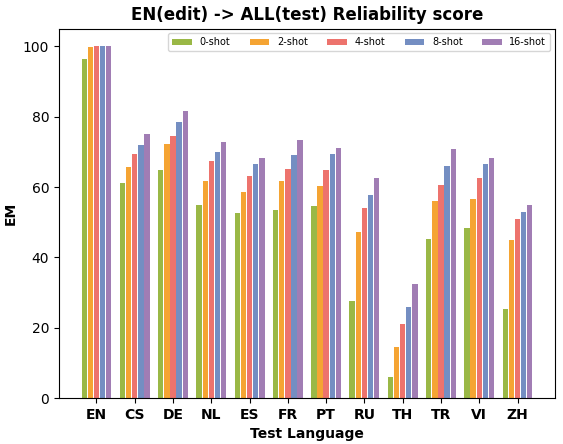}
\caption{Evaluate the performance of number of in-context example with the ReMaKE with editing in English and testing in other languages on the LLaMA backbone.}
\label{fewshot-enxx-LLaMA} 
\end{figure}

We have shown that ReMaKE-few-bi outperforms ReMaKE-zero significantly on the reliability and the generality scores. We conduct an ablation study in this subsection to understand the effect of the number of bilingual examples presented in few-shot learning on the performance of LLMs. The results of LLaMA in response to the change of the numbers of bilingual examples in a series of 2, 4, 8, 16 are illustrated in Figure~\ref{fewshot-enxx-LLaMA} for ``EN (edit) $\rightarrow$ ALL (test)''. It can be observed that ReMaKE-few-bi consistently outperforms ReMaKE-zero-LLaMA in the reliability scores, demonstrating the effects of few-shot examples in KE. The generality, locality and portability scores are shown in Figure~\ref{fewshot-enxx-LLaMA-1} in Appendix~\ref{few-appendix}. More than 16 examples would cause the problem of out-of-memory for the A100 GPU, so we set the maximum as 16.

\subsection{Ablation Study on Size of Knowledge Base}

\label{kb-size}
The above experiments are all conducted setting the knowledge base with with the whole test set. We conduct an ablation study to investigate the effect of the size of the knowledge base on a variety of benchmark metrics, including the above-mentioned four metrics (reliability, generality, locality, and portability), retrieval accuracy and time consumed. The vary the knowledge base size from 10 to 800. It can be observed in Figure~\ref{memory-size} that all benchmark metrics are slightly decreased (-0.81) with the increase in the size of the knowledge base. This is mainly caused by the minor degradation of the retriever's accuracy.


\subsection{Performance across Model Size}

\begin{figure}[htbp]
\centering          
\includegraphics[scale=0.55]{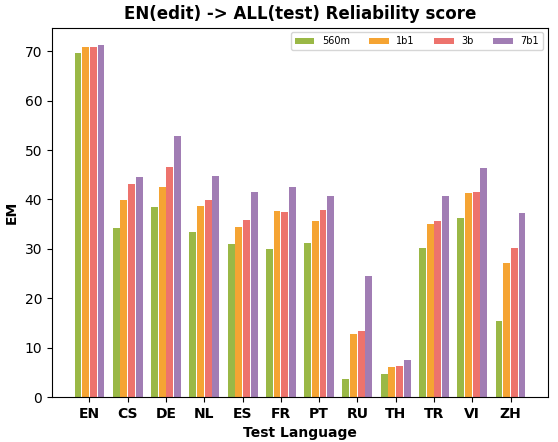}      
\caption{Evaluate the influence of model size with the ReMaKE-BLOOMZ with editing in English and testing in other languages.}
\label{model-size} 
\end{figure}

We analyze how the editing performance of ReMaKE-few-bi on the BLOOMZ backbone on ``EN (edit) $\rightarrow$ ALL (test)'' is influenced by the model size. The above-mentioned four metrics are recorded when the BLOOMZ series with a different number of parameters is selected as the backbone. Figure~\ref{model-size} and Figure~\ref{model-size-1} demonstrate a convincing win of BLOOMZ-7b over its weaker siblings in all four metrics. Even though it is hard to differentiate the performance of BLOOMZ-1b1 and BLOOMZ-3b in some specific languages, they outperform BLOOMZ-560m in all four metrics. A scale-law behavior is observed in this experiment.

\subsection{Computing Cost}

\begin{table}[htbp] \scriptsize
\begin{tabular}{p{13pt}|p{9pt}p{13pt}p{11pt}p{17pt}p{17pt}p{17pt}p{17pt}p{17pt}}
\toprule
Editor & IKE & SERAC & ROME  & \begin{tabular}[c]{@{}l@{}}ReMaKE\\ 0shot\end{tabular}  & \begin{tabular}[c]{@{}l@{}}ReMaKE\\ 4shot\end{tabular} & \begin{tabular}[c]{@{}l@{}}ReMaKE\\ 8shot\end{tabular} & \begin{tabular}[c]{@{}l@{}}ReMaKE\\ 16shot\end{tabular} \\ \hline
time   & 0.94s & 0.46s & 5.92s & \textbf{0.70s}     & 0.85s       & 1.07s    & 1.35s                   \\
\bottomrule
\end{tabular}
\caption{Time cost for each knowledge editing method conducting 1 edits on LLaMA-7b using 1X A100-80G GPU.}
\label{time-cost}
\end{table}

We gather the time consumed in KE on ``EN (edit) $\rightarrow$ ZH (test)'' in Table~\ref{time-cost} to show the computation cost efficiency for various editors. It is noted that the proposed ReMaKE-0shot achieves the best computation efficiency measured in time. The computation overhead of ReMaKE grows with the increase in the number of examples in few-shot KE.

\section{Conclusion}

In this paper, we propose ReMaKE, a retrieval-augmented multilingual knowledge editing method, to inject multilingual knowledge into LLMs by leveraging prompts composed of retrieved knowledge and user inputs. To achieve multilingual knowledge editing, we automatically construct the MzsRE dataset to cover English, Czech, German, Dutch, Spanish, French, Portuguese, Russian, Thai, Turkish, Vietnamese, and Chinese. ReMaKE is a model and language-agnostic knowledge editor not restricted to a specific LLM and language setting. Our experimental results show that ReMaKE achieves SOTA multilingual knowledge editing performance. We also share the characteristics of multilingual knowledge editing with the community to foster research along this line.

\section*{Limitation}
As we extend the initial zsRE test set to implement a multilingual knowledge base of the proposed ReMaKE, the volume of the knowledge base is limited to 743 entries. Although ReMaKE can be easily scaled up to cope with real-world applications, the implication of implementing a large-capacity knowledge base on the proposed key metrics warrants a future study. A predefined question-and-answering template is used to define multilingual knowledge contained in the knowledge base. Future work will focus on developing a formal template to accommodate a more comprehensive scope of tasks.

\bibliography{anthology,custom}
\bibliographystyle{acl_natbib}

\appendix

\section{Example Appendix}
\label{sec:appendix}

\subsection{Statistics of MzsRE} 
\label{statistics-appendix}
We list the statistics of MzsRE in 12 languages in the Table~\ref{statistics}.

\begin{table*}[h] \scriptsize
    \centering
    \begin{tabular}{cccccccc}
    \toprule
       \textbf{Lang}  & \textbf{Question} &\textbf{ \begin{tabular}[c]{@{}l@{}}Rephrased\\ Question\end{tabular}} & \textbf{Answer} & \textbf{\begin{tabular}[c]{@{}l@{}}Locality \\ Question\end{tabular}} & \textbf{\begin{tabular}[c]{@{}l@{}}Locality\\ Answer\end{tabular}}  & \textbf{\begin{tabular}[c]{@{}l@{}}Portability \\ Question\end{tabular}} & \textbf{\begin{tabular}[c]{@{}l@{}}Portability \\ Answer\end{tabular}} \\ \hline
        \textbf{EN} & 7.89 & 2.01 & 7.69 & 11.11 & 3.68 & 12.74 & 2.87 \\
        \textbf{CS} & 6.62 & 1.90 & 6.58 & 7.29 & 3.38 & 10.76 & 2.68 \\
        \textbf{DE} & 7.21 & 1.86 & 7.23 & 8.39 & 3.56 & 12.12 & 2.69 \\
        \textbf{NL} & 7.55 & 1.91 & 7.54 & 8.83 & 3.80 & 12.60 & 2.75 \\
        \textbf{ES} & 7.94 & 2.28 & 7.87 & 9.69 & 4.21 & 13.19 & 3.13 \\
        \textbf{FR} & 9.12 & 2.17 & 9.04 & 9.71 & 4.11 & 14.24 & 3.11 \\
        \textbf{PT} & 7.98 & 2.23 & 7.88 & 9.27 & 4.04 & 12.57 & 3.04 \\
        \textbf{RU} & 6.21 & 2.02 & 6.18 & 7.10 & 3.51 & 10.10 & 2.59 \\
        \textbf{TH} & 31.72 & 11.06 & 31.76 & 32.06 & 17.82 & 52.29 & 14.99 \\
        \textbf{TR} & 5.58 & 1.90 & 5.55 & 6.65 & 3.22 & 8.95 & 2.62 \\
        \textbf{VI} & 8.66 & 2.71 & 8.63 & 11.02 & 4.94 & 14.98 & 3.78 \\
        \textbf{ZH} & 19.46 & 6.05 & 19.61 & 16.90 & 9.05 & 27.16 & 7.05 \\
    \bottomrule
    \end{tabular}
    \caption{Statistics of sentence length (in word count) of MzsRE. Lang: language, EN: English, CS: Czech, DE: German, NL: Dutch, ES: Spanish, FR: French, PT: Portuguese, RU: Russian, TH: Thai, TR: Turkish, VI: Vietnamese, ZH: Chinese. }
    \label{statistics}
\end{table*}

\subsection{Results of BLOOMZ}
\label{em-bloomz}
In order to compare the results of ReMaKE with different base LLMs, we evaluate ReMaKE on the BLOOMZ, the exact match score are show in Table~\ref{en2xx-em-bloomz} and Table~\ref{xx2en-em-bloomz}.

\begin{table*}[htbp] \scriptsize
\centering
\begin{tabular}{llllllllllllll}
\toprule
\multirow{2}{*}{\textbf{Metrics}}                     & \multirow{2}{*}{\textbf{Edit on EN}}     & \multicolumn{12}{c}{\textbf{Test on}} \\ \cdashline{3-14}[1pt/1pt]
 & & \textbf{EN}     & \textbf{CS}     & \textbf{DE}     & \textbf{NL}     & \textbf{ES}     & \textbf{FR}     & \textbf{PT}     & \textbf{RU}     & \textbf{TH}     & \textbf{TR}     & \textbf{VI}     & \textbf{ZH}     \\ \hline
\multirow{3}{*}{\textbf{Reliability}} & BLOOMZ & 1.88 & 0.13	& 0.40	& 0.13		& 0.40	& 1.21	& 0.81	& 0.00	& 0.00	& 0.13	& 0.40	& 1.75 \\  \cdashline{2-14}[1pt/1pt]
& \textbf{ReMaKE}-zero & 69.04           & 29.21          & 34.59          & 28.26          & 25.03          & 27.59          & 25.98          & 0.13           & 4.44           & 21.53          & 28.80          & 18.98          \\
& \textbf{ReMaKE}-few-bi  & 71.20           & 44.55          & 52.76          & 44.68          & 41.59          & 42.53          & 40.65          & 24.50          & 7.54           & 40.65          & 46.30          & 37.28          \\ \hline
\multirow{3}{*}{\textbf{Generality}} & BLOOMZ & 1.35	& 0.13	& 0.27	& 0.13	& 0.27	& 0.81	& 0.67	& 0.00	& 0.00	& 0.13	& 0.27	& 1.88 \\  \cdashline{2-14}[1pt/1pt]
& \textbf{ReMaKE}-zero & 63.26           & 28.67          & 33.24          & 27.59          & 24.63          & 26.65          & 25.30          & 0.13           & 4.71           & 21.27          & 27.05          & 17.50          \\
& \textbf{ReMaKE}-few-bi & 65.81           & 43.47          & 51.14          & 43.07          & 39.70          & 41.32          & 39.84          & 23.28          & 7.13           & 38.63          & 44.01          & 35.94          \\\hline
\multirow{2}{*}{\textbf{Locality}}   
& \textbf{ReMaKE}-zero  & 99.19           & 98.25          & 99.60          & 99.73          & 97.85          & 98.92          & 99.19          & 97.44          & 95.29          & 97.04          & 97.44          & 94.62          \\
& \textbf{ReMaKE}-few-bi & 99.19           & 98.25          & 99.60          & 99.73          & 97.85          & 99.06          & 98.92          & 97.44          & 95.29          & 97.04          & 97.44          & 94.62          \\\hline
\multirow{3}{*}{\textbf{Portability}} & BLOOMZ & 6.59	& 0.13	& 1.35	& 0.13	& 2.29	& 2.15	& 2.15	& 0.00	& 0.00	& 0.00	& 2.29	& 4.58 \\  \cdashline{2-14}[1pt/1pt]
& \textbf{ReMaKE}-zero & 12.65           & 0.40           & 2.29           & 0.94           & 4.44           & 4.98           & 4.71           & 0.00           & 0.13           & 0.40           & 3.77           & \textbf{7.67}  \\
& \textbf{ReMaKE}-few-bi & 7.81            & 0.54           & 1.62           & 1.08           & 4.71           & 4.04           & 4.04           & 0.54           & 0.13           & 0.54           & 2.83           & 6.86           \\
\bottomrule              
\end{tabular}
\caption{\label{en2xx-em-bloomz}Exact Match (EM) results on the BLOOMZ backbone obtained from testing in English, Czech, German, Dutch, Spanish, French, Portuguese, Russian, Thai, Turkish, Vietnamese and Chinese after performing KE on knowledge in English. ``ReMaKE-few'' means the proposed knowledge editing method leveraging few-shot learning based on 16 bilingual examples concatenated in the context. ``ReMaKE-mono'' uses 16 monolingual (English) in the context. ``BLOOMZ'' are the results of pre-editing.}
\end{table*}

\begin{table*}[htbp] \scriptsize
\centering
\begin{tabular}{lllllllllllll}
\toprule

\multirow{2}{*}{\textbf{Metrics}}                     & \multirow{2}{*}{\textbf{Test on EN}}     & \multicolumn{11}{c}{\textbf{Edit on}} \\ \cdashline{3-13}[1pt/1pt]
 &    & \textbf{CS}     & \textbf{DE}     & \textbf{NL}     & \textbf{ES}     & \textbf{FR}     & \textbf{PT}     & \textbf{RU}     & \textbf{TH}     & \textbf{TR}     & \textbf{VI}     & \textbf{ZH}     \\ \hline
\multirow{3}{*}{\textbf{Reliability}} & BLOOMZ &   1.88     &  1.88       &  1.88  &  1.88     &   1.88    & 1.88     &   1.88    &   1.88   &    1.88   &   1.88     &   1.88   \\ \cdashline{2-13}[1pt/1pt]
& \textbf{ReMaKE}-zero  & 34.05          & 39.43          & 30.55          & 33.65          & 34.05          & 32.71          & 23.82          & 5.38           & 27.86          & 36.47          & 23.55          \\
& \textbf{ReMaKE}-few-bi & 48.32          & 55.99          & 49.53          & 52.89          & 48.86          & 53.43          & 36.74          & 14.00          & 46.16          & 54.91          & 45.76          \\\hline
 
\multirow{3}{*}{\textbf{Generality}} & BLOOMZ &  1.35      &   1.35      &  1.35  &   1.35    &   1.35    &   1.35   &  1.35     &  1.35    &   1.35    &   1.35     &  1.35    \\ \cdashline{2-13}[1pt/1pt]
& \textbf{ReMaKE}-zero & 32.44          & 36.34          & 28.80          & 32.57          & 32.71          & 31.36          & 21.27          & 5.11           & 25.84          & 33.38          & 21.27          \\
& \textbf{ReMaKE}-few-bi & 46.70          & 54.91          & 48.99          & 51.68          & 48.18          & 51.95          & 34.86          & 13.73          & 44.82          & 52.49          & 42.93          \\ \hline
 
\multirow{2}{*}{\textbf{Locality}}  & \textbf{ReMaKE}-zero & 98.52          & 98.38          & 98.52          & 98.38          & 98.38          & 98.52          & 97.71          & 97.17          & 96.64          & 98.52          & 98.92          \\
& \textbf{ReMaKE}-few-bi & 98.52          & 98.38          & 98.52          & 98.38          & 98.52          & 98.52          & 97.71          & 97.31          & 96.77          & 98.52          & 98.92           \\ \hline

\multirow{3}{*}{\textbf{Portability}} & BLOOMZ &   6.59     &  6.59       &  6.59  &  6.59    &   6.59    &  6.59   &   6.59    &   6.59   &  6.59     &  6.59      &   6.59   \\ \cdashline{2-13}[1pt/1pt]
& \textbf{ReMaKE}-zero & 9.02           & 9.69           & 9.69           & 9.96           & 10.90          & 11.57          & 7.67           & 6.19           & 7.54           & 9.29           & 8.88           \\
& \textbf{ReMaKE}-few-bi & 5.79           & 7.27           & 6.33           & 6.19           & 7.81           & 6.73           & 5.38           & 4.98           & 6.33           & 7.40           & 5.38           \\
\bottomrule              
\end{tabular}
\caption{\label{xx2en-em-bloomz}EM (Exact Match) results on the BLOOMZ backbone obtained from testing in English after performing KE on knowledge in Czech, German, Dutch, Spanish, French, Portuguese, Russian, Thai, Turkish, Vietnamese and Chinese. ``ReMaKE-few'' means the proposed knowledge editing method leveraging few-shot learning based on 16 bilingual examples concatenated in the context. ``ReMaKE-mono'' uses 16 monolingual (editing language) in the context. ``BLOOMZ'' are the results of pre-editing.}
\end{table*}

\subsection{English-centric F1 score}

We export the F1 score of from English to other languages and vise versa in Table~\ref{en2xx-f1} and Table~\ref{xx2en-f1}.

\begin{table*}[htbp] \scriptsize
\centering
\begin{tabular}{llllllllllllll}
\toprule
\multirow{2}{*}{\textbf{Metrics}}                     & \multirow{2}{*}{\textbf{Edit on EN}}     & \multicolumn{12}{c}{\textbf{Test on}} \\ \cdashline{3-14}[1pt/1pt]
 & &  \textbf{EN}     & \textbf{CS}     & \textbf{DE}     & \textbf{NL}     & \textbf{ES}     & \textbf{FR}     & \textbf{PT}     & \textbf{RU}     & \textbf{TH}     & \textbf{TR}     & \textbf{VI}     & \textbf{ZH}     \\ \hline
\multirow{7}{*}{\textbf{Reliability}} & \textbf{SERAC}    &  96.25           & 19.38          & 18.08          & 18.32          & 16.61          & 17.78          & 17.33          & 19.91          & 4.01           & 15.59    & 12.68 & 10.01    \\
& \textbf{IKE}          &  100.0 & 74.62          & 74.16          & 70.91          & 64.55          & 70.24          & 65.16          & 55.83          & 43.89          & 65.18   & 73.60 & 42.25  \\
& \textbf{ROME}     &  83.80  &   37.00   &  43.56    &   42.47  &   36.99  &  39.15  &  38.40  &  18.10 &   2.72   &   23.25    &  22.59    &  11.18   \\  \cdashline{2-14}[1pt/1pt]
& \textbf{ReMaKE}-zero-BLOOMZ & 89.69           & 49.41          & 57.23          & 50.17          & 52.18          & 56.39          & 52.50          & 19.91          & 23.01          & 43.75          & 58.17          & 55.64          \\
& \textbf{ReMaKE}-zero-LLaMA  & 98.05           & 79.91          & 82.43          & 75.81          & 71.99          & 75.13          & 74.13          & 67.37          & 48.91          & 69.23          & 76.28          & 68.53          \\
& \textbf{ReMaKE}-few-BLOOMZ & 91.43           & 68.58          & 74.47          & 67.88          & 69.11          & 71.44          & 70.71          & 49.47          & 46.32          & 66.92          & 72.60          & 70.08          \\
& \textbf{ReMaKE}-few-LLaMA & \textbf{100.0} & \textbf{87.61} & \textbf{90.45} & \textbf{85.99} & \textbf{82.86} & \textbf{86.99} & \textbf{85.25} & \textbf{84.28} & \textbf{69.77} & \textbf{84.38} & \textbf{86.42} & \textbf{80.23} \\ \hline
 
\multirow{7}{*}{\textbf{Generality}}  & \textbf{SERAC}        &  54.25           & 19.14          & 18.28          & 18.52          & 16.69          & 17.27          & 17.30          & 19.61          & 3.91           & 15.54   & 12.66 & 10.33    \\
& \textbf{IKE}          &  99.10           & 73.85          & 73.94          & 70.42          & 63.81          & 69.62          & 64.62          & 55.11          & 44.21          & 64.63  & 73.32 & 42.11     \\
& \textbf{ROME}     & 68.91  &  36.35   &  41.83  &   40.73  &  36.98   &  37.67  &  35.97  &  17.82  &  2.97   &  23.52   & 22.56   &  10.66   \\  \cdashline{2-14}[1pt/1pt]
& \textbf{ReMaKE}-zero-BLOOMZ & 85.02           & 48.83          & 56.05          & 49.42          & 51.28          & 55.28          & 51.53          & 19.77          & 23.30          & 43.13          & 56.00          & 53.82          \\
& \textbf{ReMaKE}-zero-LLaMA & 92.48           & 78.03          & 80.38          & 74.37          & 70.55          & 72.71          & 71.84          & 65.04          & 49.23          & 67.22          & 75.14          & 66.36          \\
& \textbf{ReMaKE}-few-BLOOMZ & 87.16           & 67.72          & 73.21          & 66.46          & 67.72          & 70.41          & 68.96          & 48.62          & 46.06          & 66.06          & 71.10          & 68.63          \\
& \textbf{ReMaKE}-few-LLaMA  & \textbf{99.07}  & \textbf{87.02} & \textbf{89.77} & \textbf{85.24} & \textbf{82.67} & \textbf{85.98} & \textbf{84.91} & \textbf{83.74} & \textbf{69.75} & \textbf{83.83} & \textbf{86.00} & \textbf{79.44} \\ \hline
 
\multirow{7}{*}{\textbf{Locality}}    & \textbf{SERAC}        & 99.80           & 100.0         & 100.0         & 100.0         & 100.0         & 100.0         & 100.0         & 100.0         & 100.0         & 100.0   &  100.0  & 99.98  \\
& \textbf{IKE}          &   67.50           & 32.71          & 38.60          & 33.96          & 34.41          & 32.94          & 33.26          & 34.88          & 53.54          & 34.04    & 41.32 & 45.65    \\
& \textbf{ROME}     &  97.83  &  95.46  &  95.99   &  95.94   &  97.20  &   96.32  &   96.20 &   95.72   &   97.53   &   95.57  &  97.91  &  97.80 \\  \cdashline{2-14}[1pt/1pt]
& \textbf{ReMaKE}-zero-BLOOMZ  & 99.50           & 98.46          & 99.63          & 99.82          & 98.55          & 99.34          & 99.40          & 97.91          & 97.38          & 97.43          & 98.14          & 96.53          \\
& \textbf{ReMaKE}-zero-LLaMA & 99.76           & 99.15          & 99.84          & 99.94          & 99.03          & 99.51          & 99.50          & 98.49          & 97.79          & 98.26          & 98.67          & 97.17          \\
& \textbf{ReMaKE}-few-BLOOMZ & 99.39           & 98.48          & 99.63          & 99.78          & 98.51          & 99.46          & 99.24          & 97.68          & 97.34          & 97.37          & 98.05          & 96.43          \\
& \textbf{ReMaKE}-few-LLaMA  & 99.76           & 98.97          & 99.71          & 99.80          & 98.71          & 99.47          & 99.47          & 98.14          & 96.73          & 98.01          & 98.49          & 96.60          \\ \hline

\multirow{7}{*}{\textbf{Portability}} & \textbf{SERAC}        &  10.06           & 2.52           & 4.65           & 4.82           & 4.44           & 4.78           & 6.11           & 4.31           & 0.74           & 1.02   & 0.47  & 0.67      \\ 
& \textbf{IKE}          &   51.96           & 35.51          & 38.48          & 36.57          & 34.74          & 37.87          & 37.23          & 39.55          & 30.60          & 28.44    & 44.83 & 23.83     \\
& \textbf{ROME}     & 9.28  &  3.10   &  5.61  &  4.73 &  4.46  & 5.02 &  5.73  &  4.32   &  0.75    &  1.13  &   0.61   &   0.73    \\  \cdashline{2-14}[1pt/1pt]
& \textbf{ReMaKE}-zero-BLOOMZ & 44.06           & 12.83          & 20.21          & 14.22          & 30.23          & 32.77          & 28.65          & 6.14           & 17.19          & 13.65          & 32.07          & 43.19          \\
& \textbf{ReMaKE}-zero-LLaMA & \textbf{64.07}  & \textbf{45.38} & \textbf{49.08} & \textbf{45.42} & \textbf{44.25} & \textbf{45.90} & \textbf{45.17} & 44.39          & 32.14          & \textbf{34.18} & \textbf{51.41} & 47.75          \\
& \textbf{ReMaKE}-few-BLOOMZ & 37.63           & 12.73          & 19.55          & 14.70          & 29.92          & 30.45          & 28.74          & 8.44           & 20.47          & 14.50          & 30.21          & 42.47          \\
& \textbf{ReMaKE}-few-LLaMA  & 62.27           & 42.89          & 44.03          & 41.84          & 41.70          & 43.41          & 42.71          & \textbf{44.64} & \textbf{33.45} & 33.37          & 47.00          & \textbf{50.23}   \\
\bottomrule              
\end{tabular}
\caption{\label{en2xx-f1}F1 results obtained from testing in Czech, German, Dutch, Spanish, French, Portuguese, Russian, Thai, Turkish, Vietnamese and Chinese after performing KE on knowledge in English. ReMaKE-few means the proposed knowledge editing method leveraging few-shot learning based on 16 bilingual examples concatenated in the context.}
\end{table*}

\begin{table*}[htbp] \scriptsize
\centering
\begin{tabular}{lllllllllllll}
\toprule
\multirow{2}{*}{\textbf{Metrics}}                     & \multirow{2}{*}{\textbf{Test on EN}}     & \multicolumn{11}{c}{\textbf{Edit on}} \\ \cdashline{3-13}[1pt/1pt]
 & &  \textbf{CS}     & \textbf{DE}     & \textbf{NL}     & \textbf{ES}     & \textbf{FR}     & \textbf{PT}     & \textbf{RU}     & \textbf{TH}     & \textbf{TR}     & \textbf{VI}     & \textbf{ZH}     \\ \hline
\multirow{6}{*}{\textbf{Reliability}} & \textbf{IKE}          &  76.58          & 75.06          & 72.94          & 64.87          & 67.62          & 66.74          & 72.03          & 4.27           & 64.38          & 60.42          & 42.94         \\
& \textbf{ROME}     &    50.18   &   66.81  &  66.78    &  61.67   &  64.70  &  65.76 &  26.88  &   4.36 &  37.05 &   27.93  &   13.00     \\  \cdashline{2-13}[1pt/1pt]
& \textbf{ReMaKE}-zero-BLOOMZ & 60.77          & 65.55          & 58.36          & 60.26          & 60.77          & 59.46          & 46.07          & 23.23          & 54.48          & 61.93          & 49.76          \\
& \textbf{ReMaKE}-zero-LLaMA & 83.71          & 81.11          & 78.45          & 78.63          & 81.67          & 77.57          & 76.91          & 33.64          & 70.14          & 71.41          & 67.56          \\
& \textbf{ReMaKE}-few-BLOOMZ & 73.49          & 78.54          & 74.50          & 76.67          & 75.91          & 77.24          & 61.49          & 35.62          & 71.65          & 77.49          & 71.07          \\
& \textbf{ReMaKE}-few-LLaMA  & \textbf{93.76} & \textbf{94.59} & \textbf{92.83} & \textbf{92.38} & \textbf{92.53} & \textbf{92.78} & \textbf{90.52} & \textbf{67.22} & \textbf{89.85} & \textbf{91.14} & \textbf{84.15} \\ \hline
 
\multirow{6}{*}{\textbf{Generality}}  & \textbf{IKE}          &  75.19          & 74.38          & 71.77          & 63.82          & 66.57          & 65.60          & 70.00          & 4.38           & 63.24          & 59.53          & 43.34     \\
& \textbf{ROME}     &   48.80  &  65.89    &  65.62      &  60.11  &   62.19  &   61.57 &  26.93  &  5.03   &  36.75  &  27.49  & 12.74   \\  \cdashline{2-13}[1pt/1pt]
& \textbf{ReMaKE}-zero-BLOOMZ & 59.11          & 63.78          & 56.71          & 58.95          & 59.88          & 58.18          & 44.22          & 22.51          & 52.40          & 60.16          & 47.51          \\
& \textbf{ReMaKE}-zero-LLaMA & 79.30          & 78.57          & 74.19          & 73.05          & 78.05          & 73.60          & 72.87          & 32.44          & 67.90          & 67.69          & 63.53          \\
& \textbf{ReMaKE}-few-BLOOMZ & 72.21          & 78.17          & 73.86          & 75.71          & 74.72          & 75.78          & 60.04          & 33.86          & 69.98          & 75.84          & 68.80          \\
& \textbf{ReMaKE}-few-LLaMA  & \textbf{93.44} & \textbf{94.00} & \textbf{92.22} & \textbf{91.23} & \textbf{91.61} & \textbf{91.70} & \textbf{89.62} & \textbf{66.81} & \textbf{88.49} & \textbf{90.70} & \textbf{82.52} \\ \hline
 
\multirow{6}{*}{\textbf{Locality}} & \textbf{IKE}          &   36.39          & 36.35          & 36.18          & 35.72          & 35.37          & 36.70          & 37.69          & 3.46           & 35.49          & 33.59          & 36.13           \\
& \textbf{ROME}     &  95.47    & 96.35   &  96.06   &  97.20   &  96.16  & 95.60  &  95.19  &   97.65  & 96.30  & 97.66  & 97.71   \\  \cdashline{2-13}[1pt/1pt]
& \textbf{ReMaKE}-zero-BLOOMZ & 98.78          & 98.93          & 98.91          & 98.93          & 98.69          & 99.01          & 98.16          & 98.00          & 97.57          & 98.87          & 99.28          \\
& \textbf{ReMaKE}-zero-LLaMA & 99.30          & 99.51          & 99.69          & 99.36          & 99.48          & 99.46          & 98.82          & 98.54          & 98.60          & 99.42          & 99.59          \\
& \textbf{ReMaKE}-few-BLOOMZ & 98.80          & 98.98          & 98.96          & 99.00          & 98.77          & 98.98          & 98.09          & 98.05          & 97.78          & 98.82          & 99.15         \\
& \textbf{ReMaKE}-few-LLaMA  & 99.09          & 99.25          & 99.34          & 99.35          & 99.18          & 99.31          & 98.78          & 98.26          & 98.44          & 99.42          & 99.46          \\ \hline

\multirow{6}{*}{\textbf{Portability}} & \textbf{IKE}          &    41.42          & 43.34          & 41.74          & 41.90          & 38.64          & 41.45          & 42.30          & 2.26           & 36.81          & 36.67          & 32.50      \\
& \textbf{ROME}     &   3.23   &   5.80  &  4.72  &  4.46   &  5.03 &  5.72 & 4.26   &  0.77 &  1.46  & 0.57 & 0.85   \\  \cdashline{2-13}[1pt/1pt]
& \textbf{ReMaKE}-zero-BLOOMZ & 38.42          & 39.80          & 39.52          & 40.49          & 41.01          & 41.40          & 34.56          & 31.81          & 37.30          & 39.35          & 38.04          \\
& \textbf{ReMaKE}-zero LLaMA & \textbf{57.57} & \textbf{59.04} & \textbf{57.45} & \textbf{57.01} & \textbf{56.89} & \textbf{57.10} & \textbf{52.87} & 41.94          & \textbf{54.61} & \textbf{55.28} & 49.87          \\
& \textbf{ReMaKE}-few-BLOOMZ & 34.30          & 35.97          & 34.83          & 35.21          & 36.70          & 35.65          & 31.39          & 31.00          & 34.71          & 34.93          & 34.34          \\
& \textbf{ReMaKE}-few-LLaMA & 52.39          & 55.25          & 54.32          & 53.68          & 54.03          & 53.75          & 48.41          & \textbf{44.70} & 51.70          & 53.67          & \textbf{49.96} \\
\bottomrule              
\end{tabular}
\caption{\label{xx2en-f1}F1 results testing in English after performing KE on knowledge in Czech, German, Dutch, Spanish, French, Portuguese, Russian, Thai, Turkish, Vietnamese and Chinese. ReMaKE-few means the proposed KE method using few-shot learning based on 16 bilingual examples concatenated in the context.}
\end{table*}

\subsection{Few Shot Learning Results}
\label{few-appendix}
\begin{figure*}[htbp]
\centering          
\subfigure[Generality of ReMaKE-LLaMA]{\label{gen-en2xx-LLaMA}\includegraphics[scale=0.55]{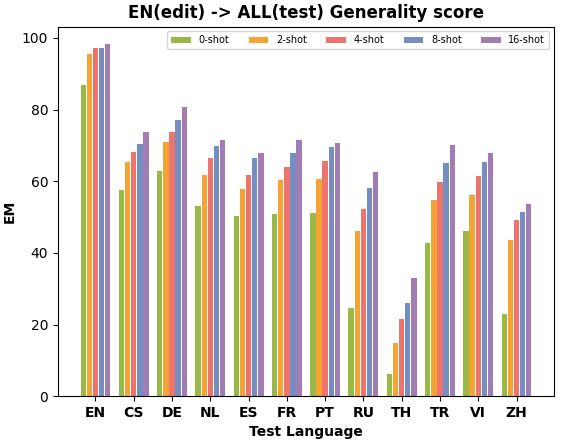}} 
\subfigure[Locality of ReMaKE-LLaMA]{\label{loc-en2xx-LLaMA}\includegraphics[scale=0.55]{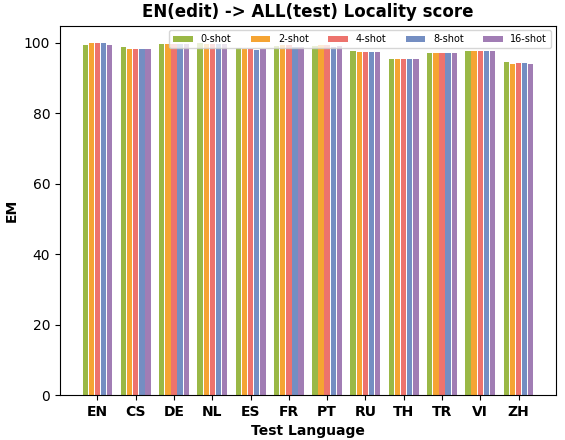}} 
\subfigure[Portability of ReMaKE-LLaMA]{\label{por-en2xx-LLaMA}\includegraphics[scale=0.55]{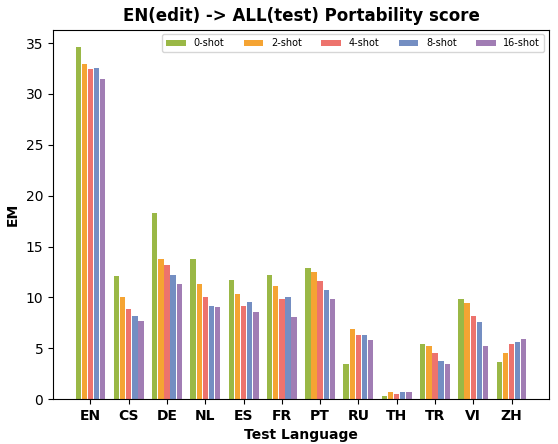}} 
\caption{Evaluate the performance of number of in-context example with the ReMaKE-LLaMA with editing in English and testing in other languages.}
\label{fewshot-enxx-LLaMA-1} 
\end{figure*}

We supplement the experimental results (Generality, Locality, Portability) of few-shot learning on ReMaKE-LLaMA in Figure~\ref{fewshot-enxx-LLaMA-1}. We reach a similar conclusion with the finding obtained in subsection ~\ref{results-portability}, in which ReMaKE-zero-LLaMA takes the lead instead in the portability score as few-shot examples tend to introduce contextual interference to the KE process. 

\subsection{Supplemental Results of Model Size}
\begin{figure*}[htbp]
\centering          
      
\subfigure[Generality of ReMaKE-BLOOMZ]{\label{scale-gen-en2xx-bloomz}\includegraphics[scale=0.55]{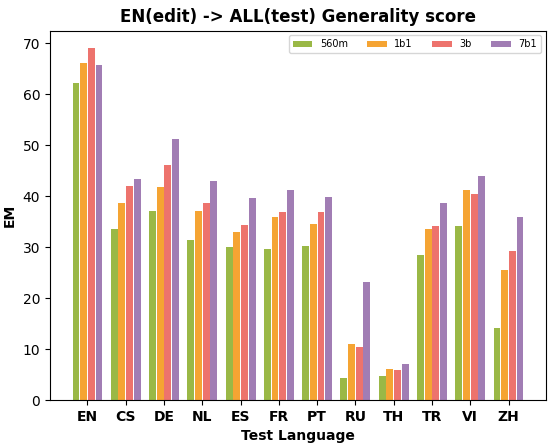}} 
\subfigure[Locality of ReMaKE-BLOOMZ]{\label{scale-loc-en2xx-bloomz}\includegraphics[scale=0.55]{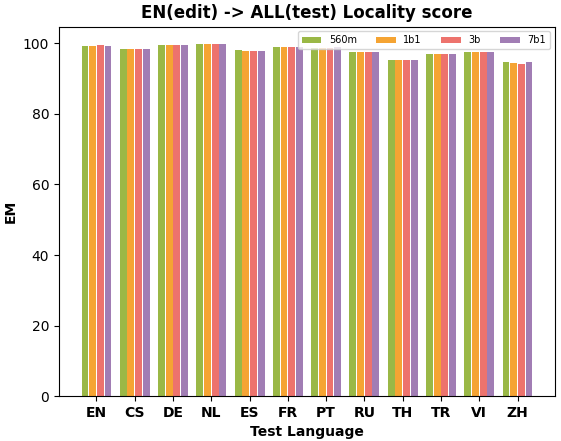}} 
\subfigure[Portability of ReMaKE-BLOOMZ]{\label{scale-por-en2xx-bloomz}\includegraphics[scale=0.55]{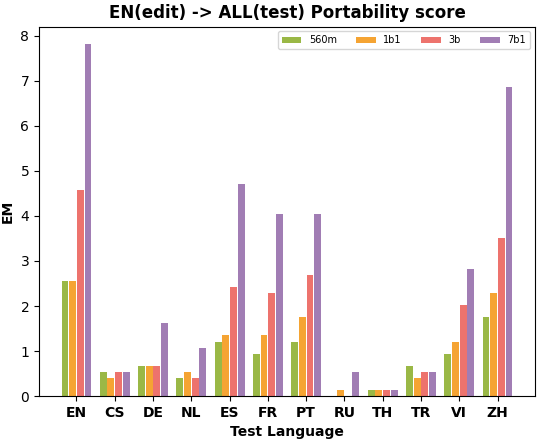}} 
\caption{Evaluate the influence of model size with the ReMaKE-BLOOMZ with editing in English and testing in other languages.}
\label{model-size-1} 
\end{figure*}

We supplement the experimental results (Generality, Locality, Portability) of different model size on ReMaKE-16shot-BLOOMZ in Figure~\ref{model-size-1}. 

\subsection{Supplemental Results Multilingual KE}
\begin{figure*}[h]
\centering             
\subfigure[Generality of ReMaKE]{\label{gen-xx2yy}\includegraphics[scale=0.25]{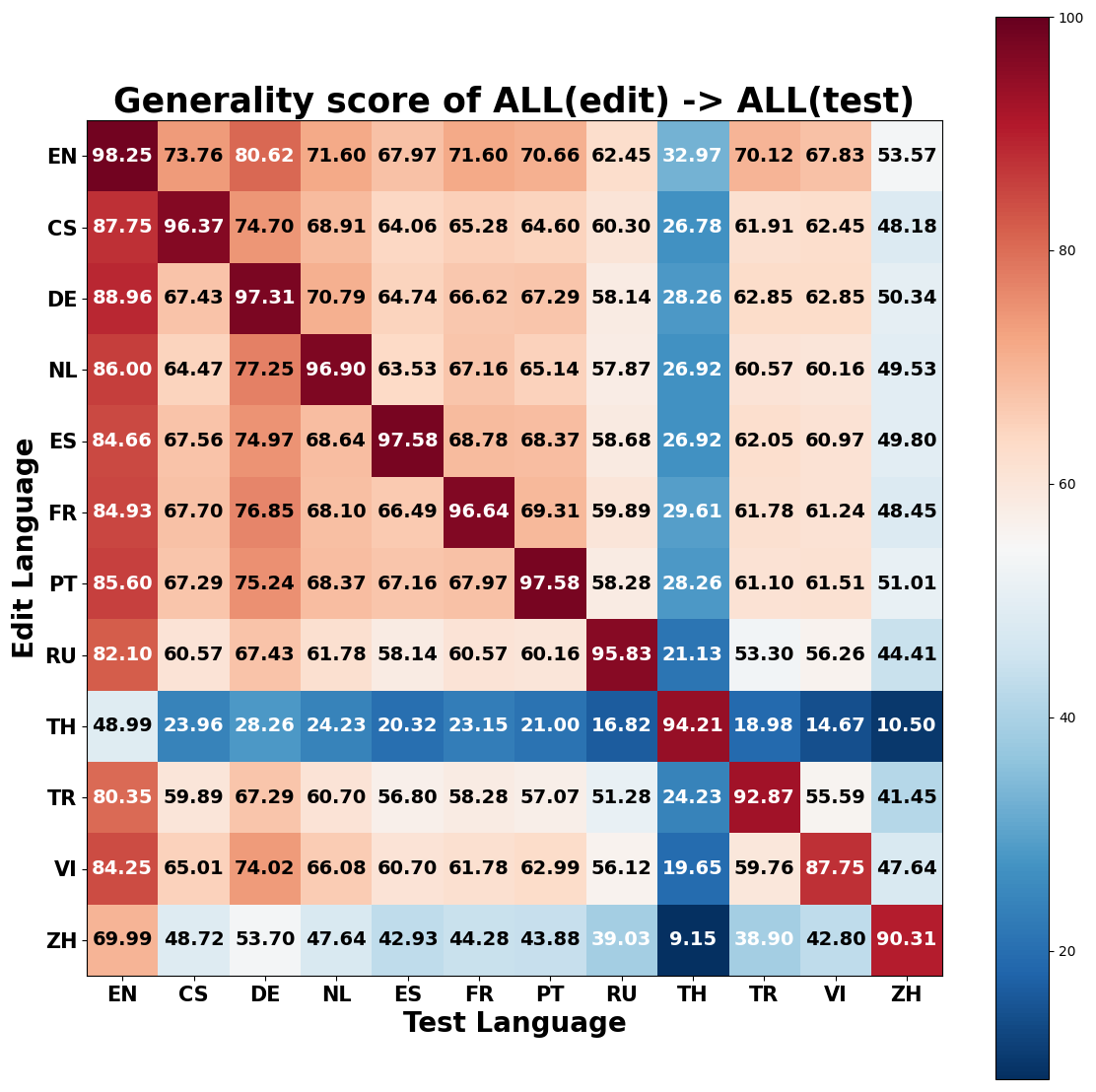}} 
\subfigure[Locality of ReMaKE]{\label{loc-xx2yy}\includegraphics[scale=0.25]{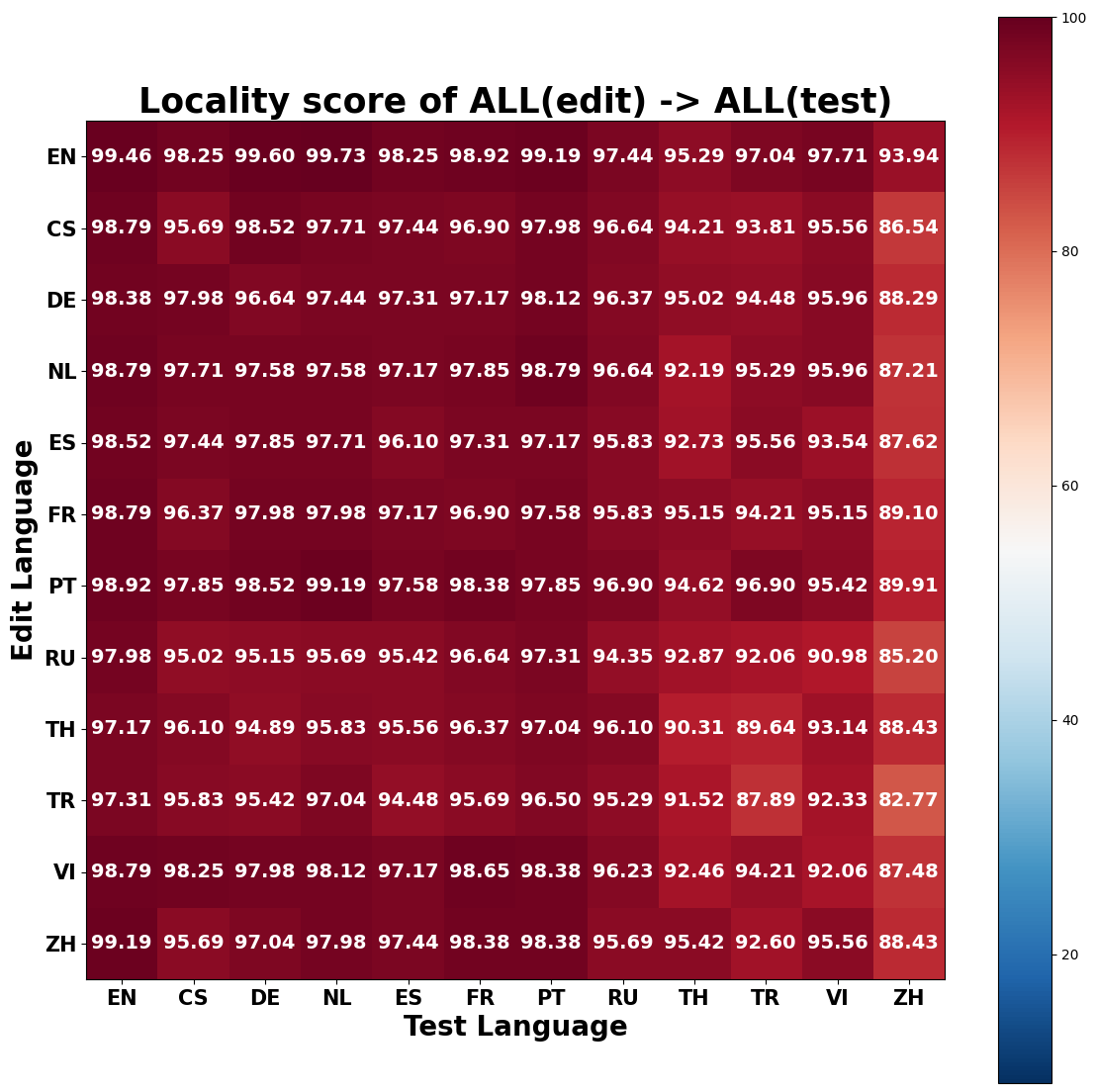}} 
\subfigure[Portability of ReMaKE]{\label{por-xx2yy}\includegraphics[scale=0.25]{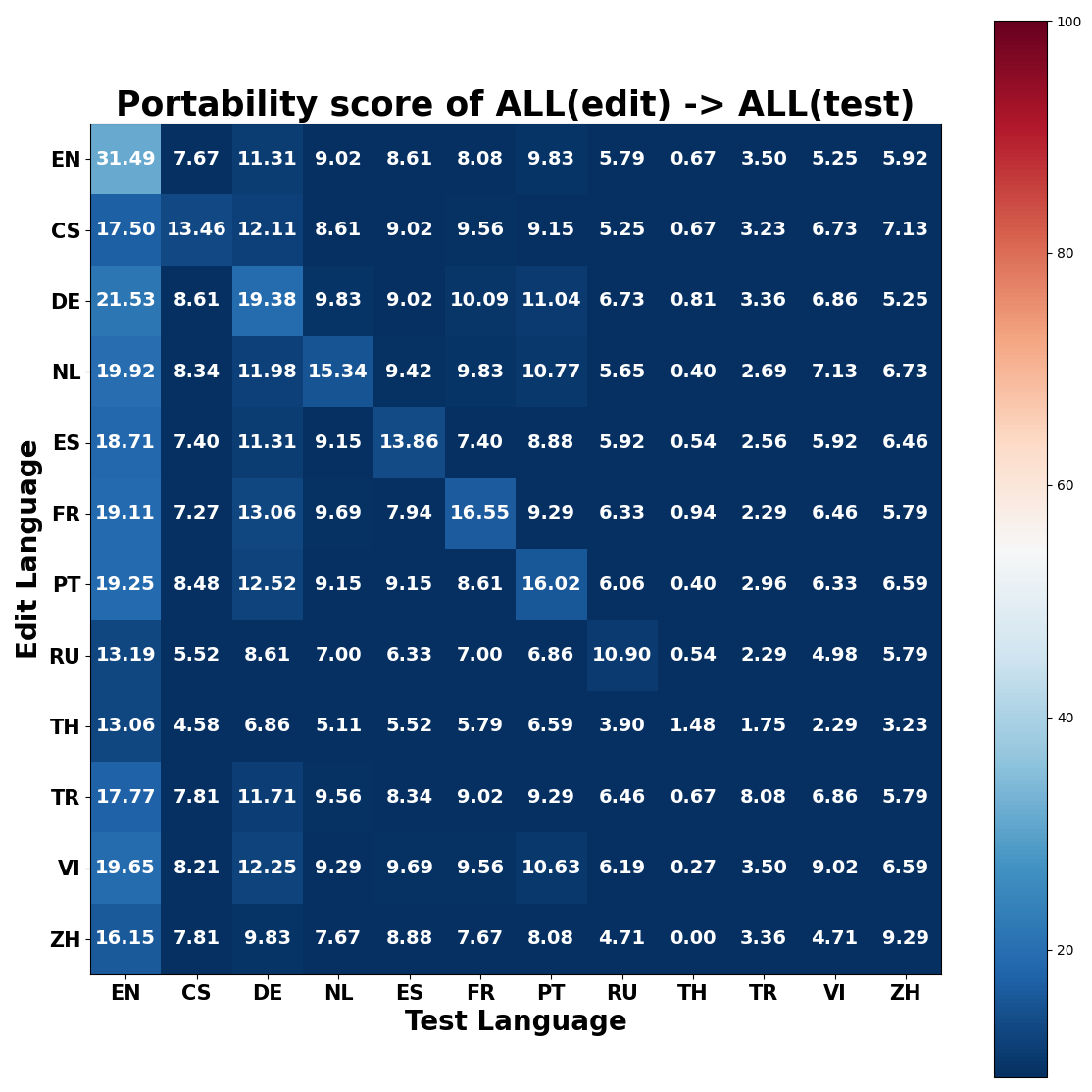}} 
\caption{Metrics based on ``ALL (edit) $\rightarrow$ ALL (test)'' editing, where ``ALL'' represents 12 languages.}
\label{acc-xx2yy-1} 
\end{figure*}

We supplement the experimental results (Generality, Locality, Portability) of multilingual knowledge editing on ReMaKE-16shot-LLaMA in Figure~\ref{acc-xx2yy-1}.

\subsection{Retriever accuracy for different test parts}

Furthermore, we evaluate the retriever for the different test set parts as shown in Table~\ref{ret-acc}. The results demonstrate that the retrieving accuracy of portability is lower than other parts, which means that the retriever lacks reasoning ability.

\begin{table*}[htbp] \scriptsize
    \centering
    \begin{tabular}{l|cccccccccccc}
\toprule
\textbf{Metrics} & \textbf{EN} & \textbf{CS} & \textbf{DE} & \textbf{NL} & \textbf{ES} & \textbf{FR} & \textbf{PT} & \textbf{RU} & \textbf{TH} & \textbf{TR} & \textbf{VI} & \textbf{ZH}   \\ \hline
\textbf{Reliability }& 100   & 100   & 99.86 & 99.87 & 99.87 & 100   & 100   & 99.46 & 99.87 & 99.87 & 99.87 & 97.85 \\
\textbf{Generality}  & 99.87 & 99.19 & 99.33 & 99.33 & 99.73 & 99.46 & 99.73 & 98.38 & 99.33 & 99.06 & 99.06 & 96.1  \\
\textbf{Locality}  & 100   & 100   & 100   & 100   & 100   & 100   & 100   & 100   & 100   & 100   & 100   & 100   \\
\textbf{Portability} & 91.79 & 88.16 & 89.23 & 89.5  & 89.64 & 89.77 & 89.23 & 81.43 & 85.6  & 89.23 & 89.23 & 84.25 \\
\bottomrule
    \end{tabular}
    \caption{Retriever accuracy for different test metrics. We evaluate retriever for the reliability, generality, locality, portability test part in MzsRE for editing in English and testing in other languages with the size of knowledge size 100.}
    \label{ret-acc}
\end{table*}

\subsection{In-context examples selection}
\label{random-sample}
\citet{few-shot} has demonstrated that the search-based prompt selection approach consistently outperforms the random selection baseline. All the above few-shot experimental results are conducted with the unsupervised prompt searching method. We compare the results of random selection and search-based strategy for examples in Table~\ref{sample}. It follows the conclusion of \citet{few-shot} that search-based selection could increase the accuracy, such as from 41.45 to 67.97 on ``EN(edit) $\rightarrow$ ES (test)''. 

\begin{table*}[htbp] \scriptsize
\centering
\begin{tabular}{lllllllllllll}
\toprule
\multirow{2}{*}{\textbf{Edit on EN}}         & \multicolumn{11}{c}{\textbf{Test on}} \\ \cdashline{2-13}[1pt/1pt]
& \textbf{EN}     & \textbf{CS}     & \textbf{DE}     & \textbf{NL}     & \textbf{ES}     & \textbf{FR}     & \textbf{PT}     & \textbf{RU}     & \textbf{TH}     & \textbf{TR}     & \textbf{VI}     & \textbf{ZH}     \\ \hline
\textbf{ReMaKE}-random-BLOOMZ &  41.32 & 35.67 & 37.15 & 30.28 & 32.17 & 36.47 &  36.34 & 3.63 &  4.98 & 31.22  & 35.4   &  26.78   \\
\textbf{ReMaKE}-search-BLOOMZ  &  70.93     &  44.41   &  52.62   &   44.55   &  41.45 &  42.40     &   40.51     &  23.82   &  7.40   &    39.97     &  45.9 & 37.01  \\
\textbf{ReMaKE}-random-LLaMA  & 54.64 & 61.91 & 79.04 & 59.89 & 55.85 & 61.91 & 61.24 & 43.88  & 8.48  &  52.22  & 55.59   &  34.05   \\
\textbf{ReMaKE}-search-LLaMA   &  \textbf{99.33} & \textbf{75.10} & \textbf{81.16} & \textbf{72.54} & \textbf{67.97} & \textbf{73.08} & \textbf{71.06} & \textbf{61.78} & \textbf{32.17} & \textbf{69.99} & \textbf{67.97} & \textbf{53.70}  \\

\bottomrule              
\end{tabular}
\caption{\label{sample} The reliability scores base on EM comparison of ReMaKE-16shot between selected examples with an unsupervised method (ReMaKE-search) and random examples (ReMaKE-random) in ``EN (edit) $\rightarrow$ ALL (test)'' editing.}
\end{table*}

\subsection{Comparison of example counts}
\begin{figure*}[htbp]
\centering          
\subfigure[Reliability of ReMaKE-BLOOMZ]{\label{rel-en2xx-bloomz}\includegraphics[scale=0.55]{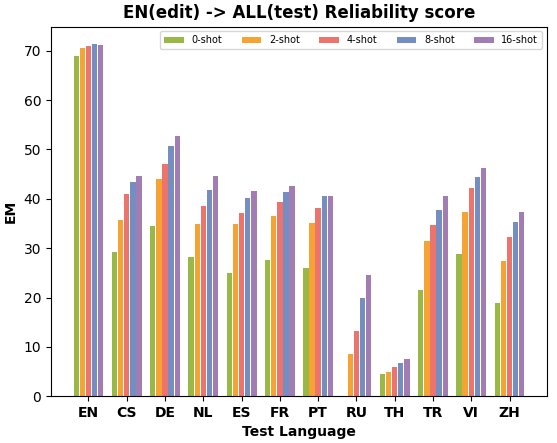}}        
\subfigure[Generality of ReMaKE-BLOOMZ]{\label{gen-en2xx-bloomz}\includegraphics[scale=0.55]{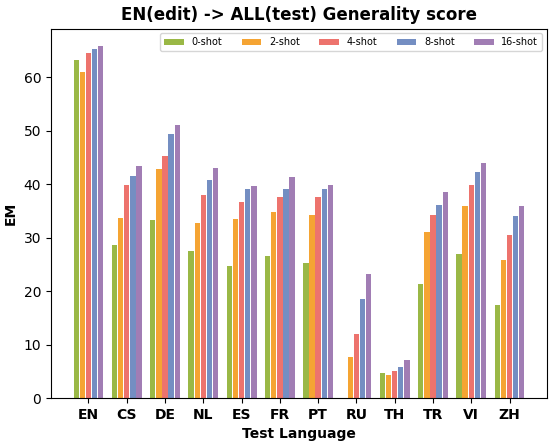}} 
\subfigure[Locality of ReMaKE-BLOOMZ]{\label{loc-en2xx-bloomz}\includegraphics[scale=0.55]{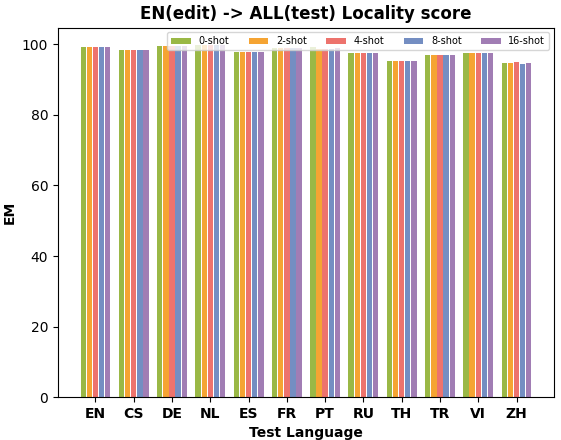}} 
\subfigure[Portability of ReMaKE-BLOOMZ]{\label{por-en2xx-bloomz}\includegraphics[scale=0.55]{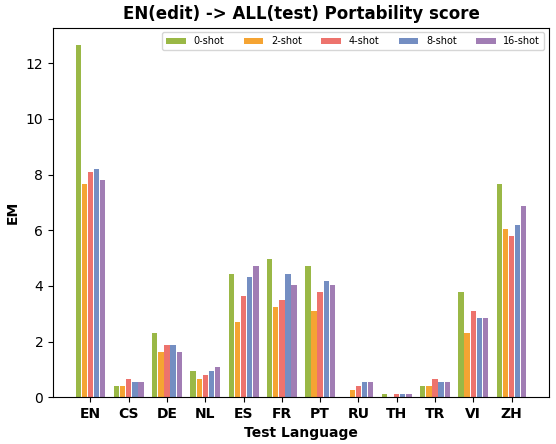}} 
\caption{Evaluate the influence of number of demonstrations with the ReMaKE-BLOOMZ with editing in English and testing in other languages.}
\label{fewshot-enxx-bloomz} 
\end{figure*}

\begin{figure*}[htbp]
\centering          
\subfigure[Reliability of ReMaKE-LLaMA]{\label{rel-xx2en-LLaMA}\includegraphics[scale=0.55]{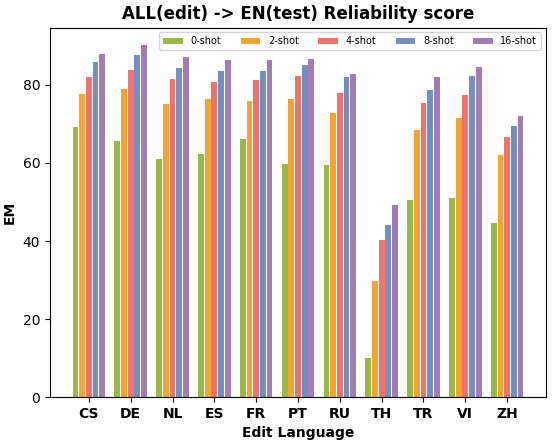}}        
\subfigure[Generality of ReMaKE-LLaMA]{\label{gen-xx2en-LLaMA}\includegraphics[scale=0.55]{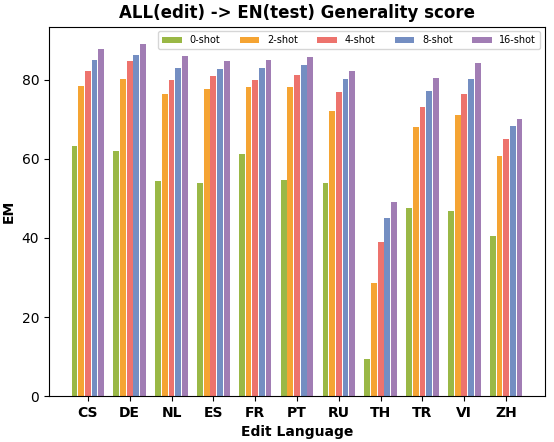}} 
\subfigure[Locality of ReMaKE-LLaMA]{\label{loc-xx2en-LLaMA}\includegraphics[scale=0.55]{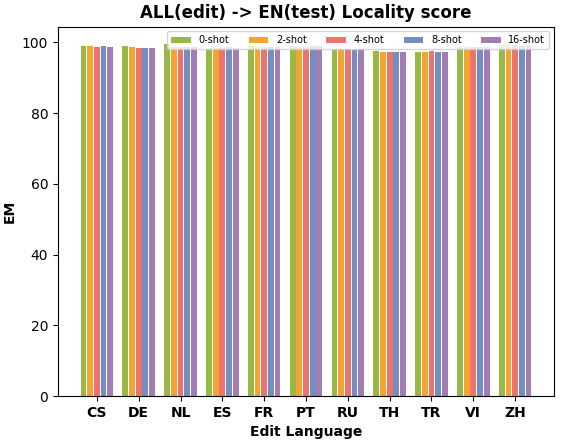}} 
\subfigure[Portability of ReMaKE-LLaMA]{\label{por-xx2en-LLaMA}\includegraphics[scale=0.55]{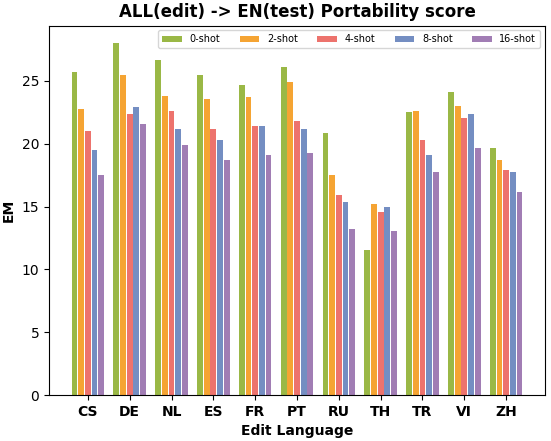}} 
\caption{Evaluate the influence of number of demonstrations with the ReMaKE-LLaMA with editing in othere languages and testing in English.}
\label{fewshot-xxen-LLaMA} 
\end{figure*}

\begin{figure*}[htbp]
\centering          
\subfigure[Reliability of ReMaKE-BLOOMZ]{\label{rel-xx2en-bloomz}\includegraphics[scale=0.55]{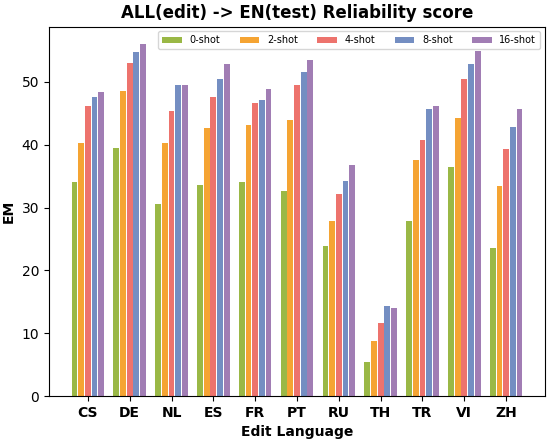}}        
\subfigure[Generality of ReMaKE-BLOOMZ]{\label{gen-xx2en-bloomz}\includegraphics[scale=0.55]{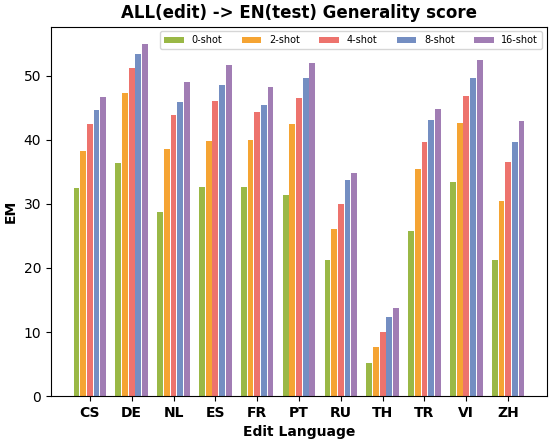}} 
\subfigure[Locality of ReMaKE-BLOOMZ]{\label{loc-xx2en-bloomz}\includegraphics[scale=0.55]{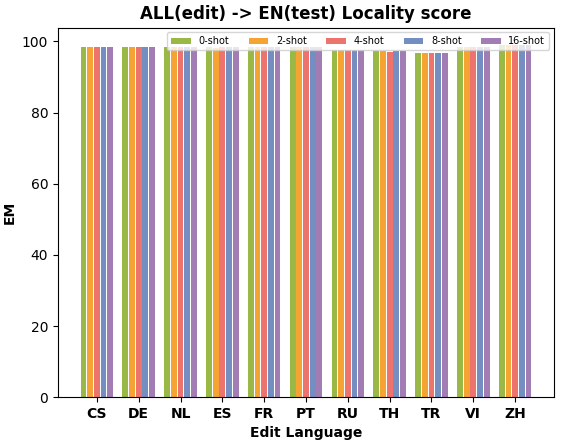}} 
\subfigure[Portability of ReMaKE-BLOOMZ]{\label{por-xx2en-bloomz}\includegraphics[scale=0.55]{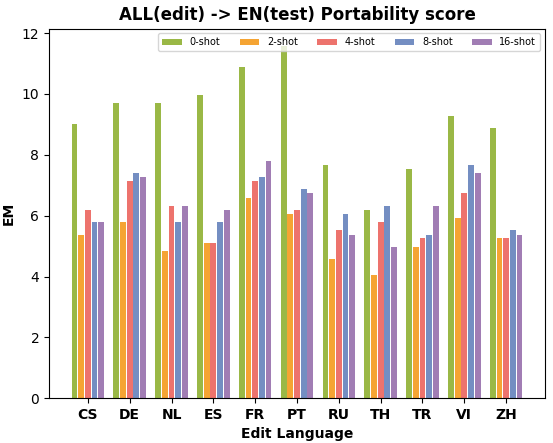}} 
\caption{Evaluate the influence of number of demonstrations with the ReMaKE-BLOOMZ with editing in othere languages and testing in English.}
\label{fewshot-xxen-bloomz} 
\end{figure*}

We supplement the comparison results of ReMaKE-BLOOMZ under the 0-shot, 2-shot, 4-shot, 8-shot, 16-shot settings from English to other languages in Table~\ref{fewshot-enxx-bloomz}. Also we conduct the same setting from other languages to English in Table~\ref{fewshot-xxen-bloomz} and Table~\ref{fewshot-xxen-LLaMA}. From the results, it proves that few-shot could greatly improve the performance compared to zero-shot for the reliability and generality.  

\end{document}